\documentclass{birkjour}

\usepackage{cleveref}
\usepackage{url}
\usepackage{amsthm}
\usepackage{amssymb}
\usepackage{bm}

\usepackage{algpseudocode,algorithm}
\algnewcommand\algorithmicinput{\textbf{Input:}}
\algnewcommand\Input{\item[\algorithmicinput]}
\algnewcommand\algorithmicoutput{\textbf{Output:}}
\algnewcommand\Output{\item[\algorithmicoutput]}

\theoremstyle{plain}
\newtheorem{theorem}{Theorem}
\newtheorem{lemma}[theorem]{Lemma}

\newtheorem{corollary}[theorem]{Corollary}

\theoremstyle{definition}
\newtheorem{definition}[theorem]{Definition}

\usepackage{tikz}
\usetikzlibrary{positioning}
\usetikzlibrary{shapes.geometric}
\usepackage{circuitikz}

\newcommand{\R}[0]{\mathbb{R}}
\newcommand{\C}[0]{\mathbb{C}}
\newcommand{\Q}[0]{\mathbb{Q}}
\newcommand{\N}[0]{\mathbb{N}}
\newcommand{\Abar}{\bar{A}}
\newcommand{\cbar}[0]{\bar{c}}
\newcommand{\Xbar}{\bar{X}}

\newcommand{\lt}{\mathrm{LT}}
\newcommand{\lm}{\mathrm{LM}}
\newcommand{\lc}{\mathrm{LC}}
\newcommand{\ideal}[1]{{\langle{#1}\rangle}}
\newcommand{\rmSigma}{\mathrm{\Sigma}}
\newcommand{\Lplus}[1]{L_{{#1}+}}
\newcommand{\Lminus}[1]{L_{{#1}-}}
\newcommand{\Lbarplus}[1]{\bar{L}_{{#1}+}}
\newcommand{\Lbarminus}[1]{\bar{L}_{{#1}-}}
\newcommand{\Splus}[1]{S_{{#1}+}}
\newcommand{\Sminus}[1]{S_{{#1}-}}

\begin{document}
    
    \title[An Inverse Kinematics Computation Using Real Quantifier Elimination]{A Design and an Implementation of an Inverse Kinematics Computation 
        in Robotics
        Using Real Quantifier Elimination 
        based on Comprehensive Gr\"obner Systems}

    \author{Shuto Otaki}
    \address{
        Graduate School of Pure and Applied Sciences \\
        University of Tsukuba \\
        Tsukuba-shi, Ibaraki 305-8571 \\
        Japan\\
        Current affiliation: \\
        Tokiwa Senior High School\\
        Ota-shi, Gunma 373-0817\\
        Japan
    }
    \email{otakishuto@math.tsukuba.ac.jp}

    \author{Akira Terui}
    \address{
        Faculty of Pure and Applied Sciences \\
        University of Tsukuba \\
        Tsukuba-shi, Ibaraki 305-8571 \\
        Japan
    }
    \email{terui@math.tsukuba.ac.jp}

    \author{Masahiko Mikawa}
    \address{
        Faculty of Library, Information and Media Science \\
        University of Tsukuba \\
        Tsukuba-shi, Ibaraki 305-8550 \\
        Japan
    }
    \email{mikawa@slis.tsukuba.ac.jp}

    \begin{abstract}
        The solution and implementation of the inverse kinematics 
        computation of a three degree-of-freedom (DOF) robot manipulator using an algorithm
        for real quantifier elimination with Comprehensive Gr\"obner Systems (CGS) are presented. 
        The method enables us to verify if the given parameters are feasible before solving the
        inverse kinematics problem. 
        Furthermore, pre-computation of CGS and substituting parameters in the CGS with the given values avoids the repetitive computation of the Gr\"obner basis.
        Experimental results compared with our previous implementation are shown.
    \end{abstract}
    

    \maketitle

    \section{Introduction}

    \subsection{The inverse kinematic problem}

    In this paper, we discuss solving the inverse kinematic problem in 
    motion planning in robotics \cite{sic-kha2008} with computer algebra. 
    In motion planning of a robot such as a manipulator with joints connected with links consecutively, we consider the forward and the inverse kinematic problems.
    While the forward kinematic problem is to determine the position of the end-effector 
    for the given configuration (angles) of the joints, 
    the inverse kinematic problem is to determine the configuration of the joints
    to bring the end-effector to the desired position.
    For the given position of the end-effector, one solves the inverse kinematic problem to obtain 
    the configuration of the joints.
    For solving the inverse kinematic problem, the forward kinematic problem is formulated,
    then the inverse kinematic problem is derived.


    Among various methods for inverse kinematics computation, methods using Gr\"obner bases 
    have been proposed 
    (\cite{fau-mer-rou2006,kal-kal1993,uch-mcp2011,uch-mcp2012},
    and the references therein). In these methods,
    the inverse kinematics problem is expressed as a system of polynomial equations with 
    trigonometric functions substituted with variables,
    along with polynomial constraints defining the relationship of these new variables.
    Then, the system of equations is 
    triangularized by computing the Gr\"obner basis with respect to the lexicographic ordering 
    and solved by appropriate solvers. 
    An advantage of the methods using Gr\"obner bases
    is that, since they solve the inverse kinematic problem directly, one can 
    verify if there exists a real solution of the inverse kinematic problem and if there  
    exists a real solution, one can obtain
    the configuration of the joints prior to the actual motion. 
    On the other hand, the computing time of Gr\"obner bases may affect the computational cost 
    of the methods: thus, it is desirable to decrease computational cost for computing 
    Gr\"obner bases and related computations.

    \subsection{Our previous work}

    Two of the present authors have proposed the implementation of inverse kinematics computation of a three degree-of-freedom (DOF) robot manipulator using Gr\"obner bases \cite{hor-ter-mik2020}. 
    The implementation uses SymPy \cite{sympy2017}, a library for computer algebra 
    written in Python, with the computer algebra system Risa/Asir
    \cite{nor2003,asir2018}, 
    connected with OpenXM infrastructure for communicating mathematical software systems 
    \cite{mae-nor-oha-tak-tam2001,openxm}. 

    We have made our implementation in this way with the following intentions. The first was for building an inverse kinematic solver with free software that is easily integrated with robotics middleware such as the Robot Operating System (ROS) \cite{ros-complete}.
    While some inverse kinematics solvers in computer algebra have been proposed using commercial computer algebra systems
    \cite{cha-mor-rou-wen2020,kaw-shi1999,net-spo1994,pit-hil-ste-max-koc2008,war2007},
    in the robotics community, much software packages are developed as free and open-source software, including ROS. Also, Python is frequently used in robotics and can easily be integrated with ROS. 
    Thus, we claim that one can easily incorporate our implementation with ROS or other software related to robotics. 
    Furthermore, for efficient computation of Gr\"obner bases with a computer algebra system 
    which can be called easily from Python and 
    distributed as free software, we have employed Risa/Asir for that purpose.

    Another one was choosing an appropriate solver for solving a system of polynomial equations among those available in Python's various packages.
    

    
    However, our implementation had its challenges: one was that 
    we were solving inverse kinematic problems without verifying the existence of a real solution. Another one was the computation of Gr\"obner basis 
    every time in solving the inverse kinematic problem.
    It would be better to avoid repetitive computation of Gr\"obner basis
    if we repeat the inverse kinematics computation many times;
    otherwise, it may make the whole computation inefficient for 
    large-scale problems such as ones with many degrees of freedom.

    \subsection{Aim of the present paper}

    In this paper, we overcome these challenges by using an algorithm for 
    real quantifier elimination based on Comprehensive Gr\"obner Systems (CGS)
    \cite{kap-sun-wan2010,mon2018,suz-sat2006},
    known as the CGS-QE algorithm, 
    which was initially proposed by Weispfenning \cite{wei1998} and improved by 
    Fukasaku et al. \cite{fuk-iwa-sat2015}.
    The inverse kinematic problem is expressed as a system of polynomial
    equations with the coordinates of the end-effector expressed as parameters.
    Then, with the CGS-QE algorithm, the existence of real roots of 
    the polynomial system is verified for the given coordinates. 
    Furthermore, by solving the system of polynomial equations defined as 
    the CGS with the parameters substituted with the given coordinates, 
    the roots of the inverse kinematic problem are computed without 
    repeated computation of Gr\"obner bases.
    Note that, although in the \emph{preprocessing steps} (see \Cref{sec:inverse-kinematics-cgsqe})
    prior to solving inverse kinematic problems, Wolfram Mathematica is used for
    simplification of formulas, the \emph{main steps} for solving 
    inverse kinematic problem is carried out with Python and Risa/Asir 
    for our intention of using free software.

    \subsection{Plan for the paper}

    This paper is organized as follows. In \Cref{sec:inverse-kinematics},
    we formulate the inverse kinematics problem of a 3 DOF manipulator for solving it using Gr\"obner bases.
    In \Cref{sec:cgs-qe}, we review a part of the CGS-QE algorithm including the 
    definition of the CGS and the theory of real root counting.
    In \Cref{sec:inverse-kinematics-cgsqe}, we present our method of solving the inverse kinematic problem using the CGS-QE algorithm.
    In \Cref{sec:exp}, we present the result of the experiments.
    Note that our implementation and the result of the experiments are freely available
    \cite{snac2021}.
    Finally, in \Cref{sec:remark}, conclusions and future research direction are presented.

    \section{Inverse kinematics of a 3 DOF robot manipulator}
    \label{sec:inverse-kinematics}

\begin{figure}[t]
        \centering
        \includegraphics[scale=0.2]{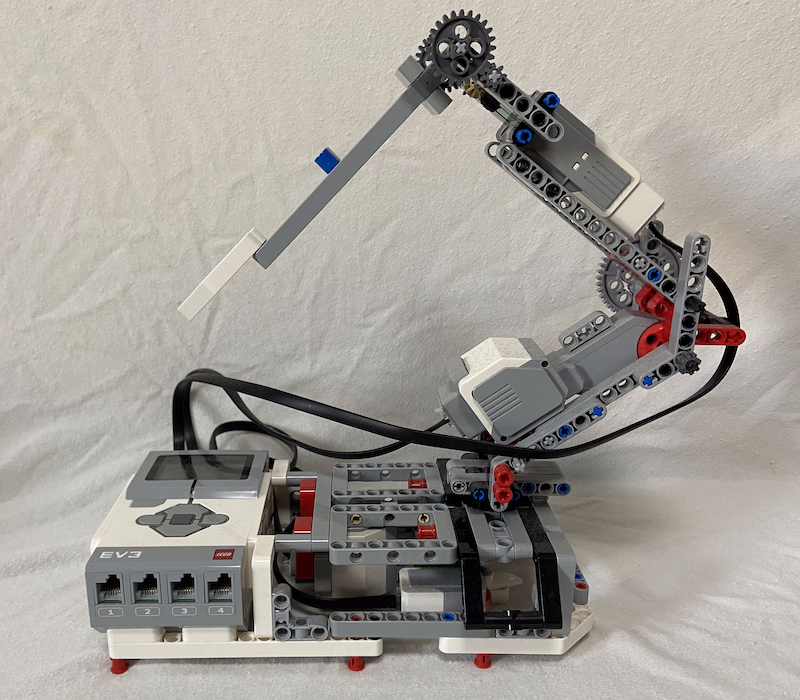}
        \caption{A 3 DOF manipulator built with EV3.}
        \label{fig:ev3-photo}
    \end{figure}

    In this paper, we consider an inverse kinematic problem of a 3 DOF robot 
    manipulator build with 
    LEGO\textsuperscript{\textregistered}
    MINDSTORMS\textsuperscript{\textregistered} EV3 Education%
    \footnote{LEGO and MINDSTORMS are trademarks of the LEGO Group.}
     (henceforth abbreviated to EV3) (\Cref{fig:ev3-photo}).
    It has a set of servo motors and sensors (gyro, ultrasonic, color, and touch sensors)
    controlled by a computer (called ``EV3 Intelligent Brick'').
    One can build a manipulator (or other kinds of robots) using bricks with these
    components, and control its motion using either a 
    GUI-based programming environment that is officially available
    or a programming language from those including Python, Ruby, C, and Java.

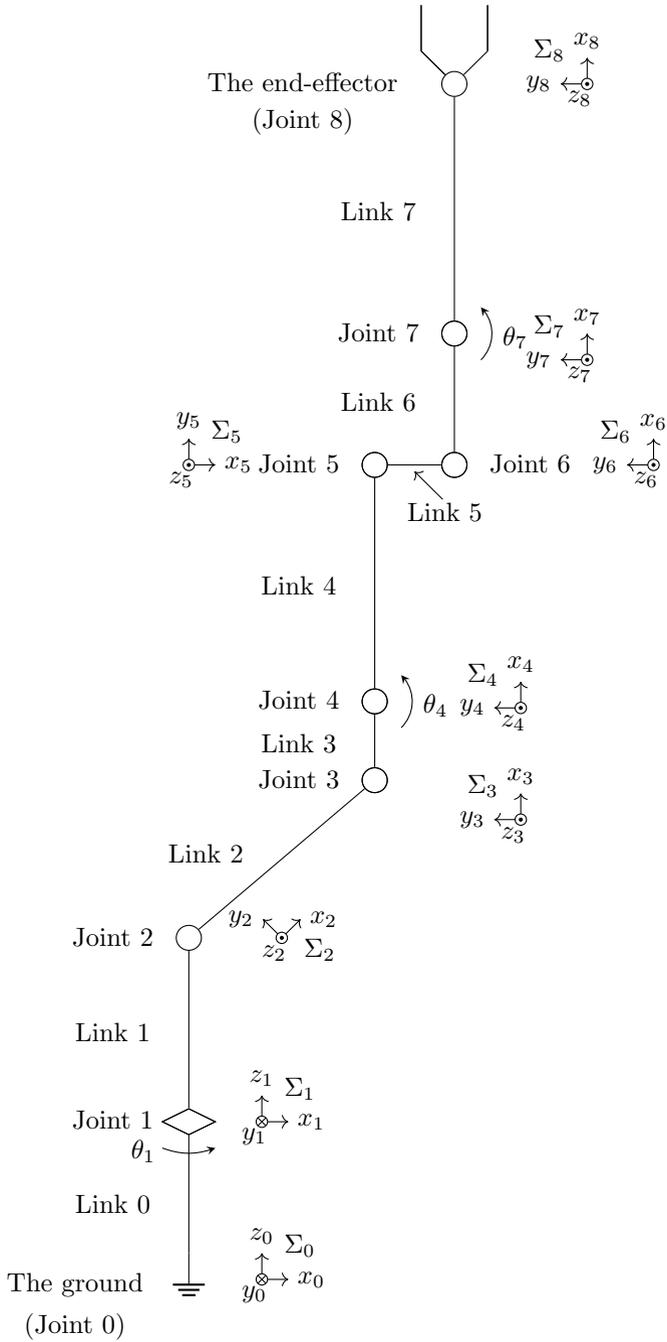
\begin{figure}
        \centering
        \begin{tikzpicture}[scale=0.175,
            cross/.style={
                path picture={ 
                    \draw[black]
                    (path picture bounding box.south east) -- 
                    (path picture bounding box.north west) 
                    (path picture bounding box.south west) -- 
                    (path picture bounding box.north east);
                }
            },
            axis/.style={
                inner sep=0pt, 
                outer sep=0pt,
                minimum width=0.15cm,
                minimum height=0.15cm
            },
            dot/.style={
                path picture={ 
                    \draw[fill, color=black]
                    circle (0.02cm);
                }
            },
            ]
    
        \node [ground,
        label={[xshift=-1.5cm, yshift=-0.7cm] The ground}](j0) at (0,1) {};
        \node [label = {[xshift=-1.5cm, yshift=-0.7cm] (Joint 0)}](j01) at (0,-3) {};
    
        \node [draw, diamond, aspect = 2,
        label={[xshift=-1cm, yshift=-0.4cm] Joint 1}](j1) at (0,11) {};
        \node [draw, diamond, aspect = 2,
        label={[xshift=-0.6cm, yshift=-0.85cm] $\theta_1$}](j1) at (0,11) {};
        \draw[->,>=stealth] (-2,9)  to [out=340,in=200] (2,9);

        \node [draw, circle,
        label={[xshift=-1cm, yshift=-0.4cm] Joint 2}](j2) at (0,25) {};
    
        \node [draw, circle](j3) at (14,37) {};
        \node [draw, circle,
        label={[xshift=-1cm, yshift=-0.4cm] Joint 3}](j3) at (14,37) {};
    
        \node [draw, circle,
        label={[xshift=0.8cm, yshift=-0.5cm] $\theta_4$}](j4) at (14,43) {};
        \node [draw, circle,
        label={[xshift=-1cm, yshift=-0.4cm] Joint 4}](j4) at (14,43) {};
        \draw[->,>=stealth] (16,41)  to [out=45,in=315] (16,45);
    
        \node [draw, circle](j5) at (14,61) {};
        \node [draw, circle,
        label={[xshift=-1cm, yshift=-0.4cm] Joint 5}](j5) at (14,61) {};
    
        \node [draw, circle](j6) at (20,61) {};
        \node [draw, circle,
        label={[xshift=1cm, yshift=-0.4cm] Joint 6}](j6) at (20,61) {};
    
        \node [draw, circle,
        label={[xshift=0.8cm, yshift=-0.5cm] $\theta_7$}](j7) at (20,71) {};
        \draw[->,>=stealth] (22,69)  to [out=45,in=315] (22,73);
        \node [draw, circle,
        label={[xshift=-1cm, yshift=-0.4cm] Joint 7}](j7) at (20,71) {};
    
        \node [draw, circle, label={[xshift=-2cm, yshift=-0.4cm] The end-effector}](j8) at (20,90) {};
        \node [label={[xshift=-2cm,yshift=-0.4cm] (Joint 8)}] (j81) at (20,87) {};
        \draw (j8)--(17.5,92.5);
        \draw (17.5,92.5)--(17.5,96);
        \draw (22.5,92.5)--(22.5,96);
    
        \draw (j1) -- (j0)
        node [midway, label={[xshift=-1cm,yshift=-0.5cm] Link 0}] {};;
    
        \draw (j1) -- (j2)
        node [midway, label={[xshift=-1cm,yshift=-0.4cm] Link 1}] {};;
        
        \draw (j2) -- (j3)
        node [midway, label={[xshift=-1cm,yshift=-0.3cm] Link 2}] {};;
    
        \draw (j3) -- (j4)
        node [midway, label={[xshift=-1cm,yshift=-0.4cm] Link 3}] {};;
    
        \draw (j4) -- (j5)
        node [midway, label={[xshift=-1cm,yshift=-0.4cm] Link 4}] {};;
    
        \draw (j5) -- (j6)
        node [midway, label={[xshift=0.4cm,yshift=-1cm] Link 5}] {};;
        \draw [<-] (17,60.5) -- ++(-45:3cm);
    
        \draw (j6) -- (j7)
        node [midway, label={[xshift=-1cm,yshift=-0.4cm] Link 6}] {};;
    
        \draw (j7) -- (j8)
        node [midway, label={[xshift=-1cm,yshift=-0.4cm] Link 7}] {};;
        \draw (j8)--(17.5,92.5);
        \draw (j8)--(22.5,92.5);
        \draw (17.5,92.5)--(17.5,96);
        \draw (22.5,92.5)--(22.5,96);

        \node [
            draw, circle, axis, cross,
            label={[xshift=-0.1cm, yshift=-0.5cm] $y_0$},
            label={[xshift=0.5cm, yshift=0.1cm] $\rmSigma_0$}
        ](s0) at (5.5,-1){};
        \draw [->] (s0) -- ++(90:2cm) node [above] {$z_0$};
        \draw [->] (s0) -- ++(0:2cm) node [right] {$x_0$};
    
        \node [
            draw, circle, axis, cross,
            label={[xshift=-0.1cm, yshift=-0.5cm] $y_1$},
            label={[xshift=0.5cm, yshift=0.1cm] $\rmSigma_1$}
        ](s0) at (5.5,11){}; 
        \draw [->] (s0) -- ++(90:2cm) node [above] {$z_1$};
        \draw [->] (s0) -- ++(0:2cm) node [right] {$x_1$};
    
         \node [
            draw, circle, axis, dot,
            label={[xshift=-0.1cm, yshift=-0.5cm] $z_2$},
            label={[xshift=0.5cm, yshift=-0.5cm] $\rmSigma_2$}
        ](s2) at (7,25){};
        \draw [->] (s2) -- ++(45:2cm) node [right] {$x_2$};
        \draw [->] (s2) -- ++(135:2cm) node [left] {$y_2$};
    
        \node [
            draw, circle, axis, dot,
            label={[xshift=-0.1cm, yshift=-0.5cm] $z_3$},
            label={[xshift=-0.5cm, yshift=0.1cm] $\rmSigma_3$}
        ](s0) at (25,34){};
        \draw [->] (s0) -- ++(90:2cm) node [above] {$x_3$};
        \draw [->] (s0) -- ++(180:2cm) node [left] {$y_3$};
    
         \node [
            draw, circle, axis, dot,
            label={[xshift=-0.1cm, yshift=-0.5cm] $z_4$},
            label={[xshift=-0.5cm, yshift=0.1cm] $\rmSigma_4$}
        ](s0) at (25,42.5){};
        \draw [->] (s0) -- ++(90:2cm) node [above] {$x_4$};
        \draw [->] (s0) -- ++(180:2cm) node [left] {$y_4$};
    
        \node [
            draw, circle, axis, dot,
            label={[xshift=-0.1cm, yshift=-0.5cm] $z_5$},
            label={[xshift=0.5cm, yshift=0.1cm] $\rmSigma_5$}
        ](s0) at (0,61){};
        \draw [->] (s0) -- ++(90:2cm) node [above] {$y_5$};
        \draw [->] (s0) -- ++(0:2cm) node [right] {$x_5$};
    
        \node [
            draw, circle, axis, dot,
            label={[xshift=-0.1cm, yshift=-0.5cm] $z_6$},
            label={[xshift=-0.5cm, yshift=0.1cm] $\rmSigma_6$}
        ](s0) at (35,61){};
        \draw [->] (s0) -- ++(90:2cm) node [above] {$x_6$};
        \draw [->] (s0) -- ++(180:2cm) node [left] {$y_6$};
    
        \node [
            draw, circle, axis, dot,
            label={[xshift=-0.1cm, yshift=-0.5cm] $z_7$},
            label={[xshift=-0.5cm, yshift=0.1cm] $\rmSigma_7$}
        ](s0) at (30,69){};
        \draw [->] (s0) -- ++(90:2cm) node [above] {$x_7$};
        \draw [->] (s0) -- ++(180:2cm) node [left] {$y_7$};
    
        \node [
            draw, circle, axis, dot,
            label={[xshift=-0.1cm, yshift=-0.5cm] $z_8$},
            label={[xshift=-0.5cm, yshift=0.1cm] $\rmSigma_8$}
        ](s0) at (30,90){};
        \draw [->] (s0) -- ++(90:2cm) node [above] {$x_8$};
        \draw [->] (s0) -- ++(180:2cm) node [left] {$y_8$};

        \end{tikzpicture}
        \caption{Components and the coordinate systems of the manipulator.}
        \label{fig:ev3-components}
    \end{figure}

    The components of the manipulator are shown as in 
    \Cref{fig:ev3-components}\footnote{While the manipulator itself is the same one used in our previous paper \cite{hor-ter-mik2020}, we have re-measured the structure of joints and links; thus, the derived inverse kinematic problem in the present paper is slightly different 
    from the one in the previous paper.}. 
    Links 
    are called Link $i$ ($i=0,\dots,7$) from the one fixed on the ground 
    towards the end-effector.
    For $j=1,\dots,7$,
    a joint connecting Link $j-1$ and $j$ is 
    called Joint $j$. Note that Joints $1,4,7$ are revolute joints, while the other 
    joints are fixed, and all the joints and links are located on a plane.
    Among Joints $1,4,7$, Joint 1 has a diamond shape because the axis of rotation
    overlaps with Links $0$ and $1$, while Joints 4 and 7 are drawn in a circle
    because the axis of rotation is perpendicular to the links connected to the joint.

    For Joint $j$ with $j=1,\dots,7$, the coordinate system $\rmSigma_j$, with the $x_j$, $y_j$
    and $z_j$ axes and the origin at Joint $j$, is defined according to 
    a modified Denavit–Hartenberg convention \cite{wal-sch2008}
    (\Cref{fig:ev3-components}), as follows.
    \begin{itemize}
        \item The $z_j$ axis is chosen along with the axis of Joint $j$.
        \item The $x_{j-1}$ axis is chosen along with with the common normal to axes $z_{j-1}$ to
        $z_j$.
        \item The $y_j$ axis is chosen so that the present coordinate system is right-handed.
    \end{itemize}
    Note that, since the present coordinate system is right-handed, the positive axis pointing upwards and downwards are denoted by ``$\odot$'' and ``$\otimes$'', respectively.
    Now,
    let us regard the perpendicular foot on the ground from Joint 1 as Joint 0 and 
    let $\rmSigma_0$ be the coordinate system with the origin placed at the position of Joint 0,
    where the direction of axes $x_0$, $y_0$, and $z_0$ are the same as that of 
    axes $x_1$, $y_1$ and $z_1$, respectively. 
    Also, let us regard the end-effector as Joint 8 and let $\rmSigma_8$ be the coordinate system with the origin placed at the position of Joint 8,
    where the direction of axes $x_8$, $y_8$, and $z_8$ are the same as that of 
    axes $x_7$, $y_7$, and $z_7$, respectively.

    For analyzing the motion of the manipulator, we define a map between the \emph{joint space}
    and the \emph{configuration space} or \emph{operational space}. For a joint space, since we have revolute joints 
    $1,4,7$, their angles $\theta_1,\theta_4,\theta_7$, respectively, are located 
    in a circle $S^1$, we define the joint space as $\mathcal{J}=S^1\times S^1\times S^1$.
    For a configuration space, let $(x,y,z)$ be the position of the end-effector located in 
    $\R^3$ and then define the configuration space as $\mathcal{C}=\R^3$. Thus, we consider a map 
    $f:\mathcal{J}\longrightarrow\mathcal{C}$. The forward kinematic problem is to find 
    the position of the end-effector in $\mathcal{C}$ for the given configuration of 
    the joints in $\mathcal{J}$, while the inverse kinematic problem is 
    to find the configuration of the joints in $\mathcal{J}$ which enables the 
    given position of the end-effector in $\mathcal{C}$. 
    We first solve the forward kinematic problem for formulating the inverse kinematic problem. 

    Let $a_j$ be the length of the common perpendicular line segment
    of axes $z_{j-1}$ and $z_j$,
    $\alpha_j$ the angle between axes $z_{j-1}$ and $z_j$ with respect to the
    $x_{j-1}$ axis, 
    $d_j$ the length of the common perpendicular line segment
    of axes $x_{j-1}$ and $x_j$, 
    and $\theta_j$ be the angle between axes $x_{j-1}$ and $x_j$
    with respect to $z_j$ axis. Then, the coordinate transformation matrix
    ${}^{j-1}T_j$ from the coordinate system $\rmSigma_i$ to $\rmSigma_{j-1}$
    is expressed as
    \begin{align*}
        {}^{i-1} T_i &=
        \begin{pmatrix}
            1 & 0 & 0 & a_j \\
            0 & 1 & 0 & 0 \\
            0 & 0 & 1 & 0 \\
            0 & 0 & 0 & 1
        \end{pmatrix}
        \times
        \begin{pmatrix}
            1 & 0 & 0 & 0 \\
            0 & \cos \alpha_j & - \sin \alpha_j & 0  \\
            0 & \sin \alpha_j & \cos \alpha_j & 0 \\
            0 & 0 & 0 & 1
        \end{pmatrix}
        \\
        &\quad \times
        \begin{pmatrix}
            1 & 0 & 0 & 0 \\
            0 & 1 & 0 & 0 \\
            0 & 0 & 1 & d_j \\
            0 & 0 & 0 & 1
        \end{pmatrix}
        \times
        \begin{pmatrix}
            \cos \theta_j & -\sin \theta_j & 0 & 0 \\
            \sin \theta_j & \cos \theta_j & 0 & 0 \\
            0 & 0 & 1 & 0 \\
            0 & 0 & 0 & 1 
        \end{pmatrix}
        \\
        &= 
        \begin{pmatrix}
            \cos \theta_j & -\sin \theta_j & 0 & a_j \\
            \cos \alpha_j \sin \theta_j & \cos \alpha_j \cos \theta_j & -\sin \alpha_j & -d_j \sin \alpha_j   \\
            \sin \alpha_j \sin \theta_j & \sin \alpha_j \cos \theta_j & \cos \alpha_j & d_j \cos \alpha_j \\
            0 & 0 & 0 & 1
        \end{pmatrix}
        ,
    \end{align*}
    where the joint parameters $a_j$, $\alpha_j$, $d_j$ and $\theta_j$ 
    are given as shown in \Cref{tab:ev3-parameters} (note that the unit
    of $a_j$ and $d_j$ is millimeters).
    \begin{table}[t]
        \caption{Joint parameters for EV3.}
        \label{tab:ev3-parameters}
        \begin{center}
        \begin{tabular}{c|cccc}
                $j$ & $a_j$ (mm) & $\alpha_j$ & $d_j$ (mm) & $\theta_j$  \\ 
                \hline
                1 & 0 & 0 & 80 & $\theta_1 $ \\
                2 & 0 & $\pi/2$ & 0 & $\pi/4$ \\
                3 & ${88}$ & 0 & 0 &$\pi/4 $ \\
                4 & ${24} $ & 0 & 0 & $\theta_4$\\
                5 & ${96} $ & 0 & 0 & $-\pi/2$ \\
                6 & ${16}$ & 0 & 0 & $\pi/2$\\
                7 & ${40} $ & 0 & 0 &  $\theta_7$\\  
                8 & ${112} $ & 0 & 0 &0  \\
                \hline
        \end{tabular}
        \end{center}
    \end{table}
    The transformation matrix $T$ 
    from the coordinate system $\rmSigma_8$ to $\rmSigma_0$
    is calculated as 
    $T={}^{0}T_1{}^{1}T_2{}^{2}T_3{}^{3}T_4{}^{4}T_5{}^{5}T_6{}^{6}T_7{}^{7}T_8$,
    where 
    \begin{align}
        \label{siki10}
        {}^{0}T_1  
        &= 
        \left(
        \begin{array}{cccc}
        \cos \theta_1 & -\sin \theta_1 & 0 & 0 \\
        \sin \theta_1 & \cos \theta_1 & 0 & 0  \\
        0& 0 & 1 & 80 \\
        0 & 0 & 0 & 1
        \end{array}
        \right) ,&
        {}^{1}T_2
        &=
        \left(
        \begin{array}{cccc}
        \frac{\sqrt{2}}{2}  & -\frac{\sqrt{2}}{2}  & 0 & 0 \\
        0 & 0 & -1 & 0 \\
        \frac{\sqrt{2}}{2} & \frac{\sqrt{2}}{2} & 0 & 0  \\
        0 & 0 & 0 & 1
        \end{array}
        \right) ,  \notag\\
        {}^{2}T_3 
        &= 
        \left(
        \begin{array}{cccc}
        \frac{\sqrt{2}}{2} & -\frac{\sqrt{2}}{2} & 0 & {88} \\
        \frac{\sqrt{2}}{2} & \frac{\sqrt{2}}{2} & 0 & 0  \\
        0& 0 & 1 & 0 \\
        0 & 0 & 0 & 1
        \end{array}
        \right) ,&
        {}^{3}T_4
        &= 
        \left(
        \begin{array}{cccc}
        \cos \theta_4 & -\sin \theta_4 & 0 & {24} \\
        \sin \theta_4 & \cos \theta_4 & 0 & 0  \\
        0& 0 & 1 & 0 \\
        0 & 0 & 0 & 1
        \end{array}
        \right) , \notag\\
        {}^{4}T_5
        &= 
        \left(
        \begin{array}{cccc}
        0 & 1 & 0 & {96} \\
        -1 & 0 & 0 & 0 \\
        0& 0 & 1 & 0 \\
        0 & 0 & 0 & 1
        \end{array}
        \right) ,&
        {}^{5}T_6  
        &= 
        \left(
        \begin{array}{cccc}
        0 & -1 & 0 & {16} \\
        1 & 0 & 0 & 0  \\
        0& 0 & 1 & 0 \\
        0 & 0 & 0 & 1
        \end{array}
        \right) , \notag\\
        {}^{6}T_7
        &= 
        \left(
        \begin{array}{cccc}
         \cos \theta_7 & -\sin \theta_7 & 0 & {40}  \\
         \sin \theta_7 & \cos \theta_7 & 0 & 0 \\
        0& 0 & 1 & 0 \\
        0 & 0 & 0 & 1
        \end{array}
        \right) ,&
        {}^{7}T_8  
        &= 
        \left(
        \begin{array}{cccc}
        1 & 0 & 0 & {112}  \\
        0 & 1 & 0 & 0  \\
        0& 0 & 1 & 0 \\
        0 & 0 & 0 & 1
        \end{array}
        \right) . \\ \notag
    \end{align}

    Then, the position $(x,y,z)$ 
    of the end-effector with respect to the coordinate system $\rmSigma_0$ is
    expressed as 
    \begin{equation}
        \label{eq:ev3-forward-kinematics}
        \begin{split}
            x &= -112\cos{\theta_1}\cos{\theta_4}\sin{\theta_7}+16\cos{\theta_1}\cos{\theta_4}-112\cos{\theta_1}\sin{\theta_4}\cos{\theta_7}\\
            &\qquad -136{\cos}\theta_1{\sin}\theta_4+44{\sqrt{2}}\cos{\theta_1},  \\
            y &= -112\sin{\theta_1}\cos{\theta_4}\sin{\theta_7}+16\sin{\theta_1}\cos{\theta_4}-112\sin{\theta_1}\sin{\theta_4}\cos{\theta_7}\\
            &\qquad -136\sin{\theta_1}\sin{\theta_4}+44{\sqrt{2}}\sin{\theta_1},  \\
            z &= 112\cos{\theta_4}\cos{\theta_7}+136{\cos}{\theta_4}-112\sin{\theta_4}\sin{\theta_7}+16{\sin}{\theta_4}\\
            &\qquad +104+44{\sqrt{2}}.
         \end{split}
    \end{equation}

    The inverse kinematic problem is solving \Cref{eq:ev3-forward-kinematics} with respect to 
    $\theta_1$, $\theta_4$, $\theta_7$. 
    By substituting trigonometric functions $\cos\theta_j$ and $\sin\theta_j$ with variables as
    \[
    c_j=\cos\theta_j,\quad s_j=\sin\theta_j,
    \]
    subject to $c_j^2+s_j^2=1$, \Cref{eq:ev3-forward-kinematics} is transferred to a system of 
    polynomial equations:
    \begin{equation}
        \label{eq:inverse-kinematic-equations}
        \begin{split}
            f_1 &= 112c_1c_4s_7-16c_1c_4+112c_1s_4c_7+136c_1s_4-44{\sqrt{2}}c_1 + x = 0, \\
            f_2 &= 112s_1c_4s_7-16s_1c_4+112s_1s_4c_7+136s_1s_4-44{\sqrt{2}}s_1 + y = 0, \\
            f_3 &= -112c_4c_7-136c_4+112s_4s_7-16s_4-104-44{\sqrt{2}} + z = 0, \\
            f_4 &= s_1^2+c_1^2-1=0, \quad
            f_5 = s_4^2+c_4^2-1=0, \quad
            f_6 = s_7^2+c_7^2-1=0.
        \end{split}
    \end{equation}

    \section{Real quantifier elimination based on CGS}
    \label{sec:cgs-qe}

    \Cref{eq:inverse-kinematic-equations} shows that solving the inverse kinematic problem 
    for the given system can be regarded as a real quantifier elimination of 
    a quantified formula
    \begin{multline}
        \label{eq:inverse-kinematic-qe-formula}
        \exists{c_1}\exists{s_1}\exists{c_4}\exists{s_4}\exists{c_7}
        \exists{s_7} \\
        (f_1=0 \land f_2=0 \land f_3=0 \land f_4=0 \land f_5=0 \land f_6=0),
    \end{multline}
    with $x,y,z$ as parameters.

    In this section, we briefly review an algorithm of real quantifier elimination 
    based on CGS, the CGS-QE algorithm,
    by Fukasaku et al. \cite{fuk-iwa-sat2015}.
    Two main tools play a crucial role in the algorithm: one is CGS, 
    and another is real root counting, or counting the number of real roots of a system of polynomial equations.
    Note that, in this paper, we only consider equations in the quantified formula. 

    Hereafter, let $R$ be a real closed field, $C$ be the algebraic closure of $R$, and
    $K$ be a computable subfield of $R$. 
    In this paper, we consider $R$ as the field of 
    real numbers $\R$, $C$ as the field of complex numbers $\C$, and $K$ as 
    the field of rational numbers $\Q$.
    Let $\Xbar$ and $\Abar$ denote variables $X_1,\dots,X_n$ and $A_1,\dots,A_m$,
    respectively, and $T(\Xbar)$ be the set of the monomials
    which consist of variables in $\Xbar$. For an ideal $I\subset K[\Xbar]$, let
    $V_R(I)$ and $V_C(I)$ be the affine varieties of $I$ in $R$ or $C$, respectively, 
    satisfying that 
    $V_R(I)=\{\cbar\in R^n\mid \mbox{$\forall f(\Xbar)\in I$: $f(\cbar)=0$}\}$
    and
    $V_C(I)=\{\cbar\in C^n\mid \mbox{$\forall f(\Xbar)\in I$: $f(\cbar)=0$}\}$.

    \subsection{CGS}

    For the detail and algorithms on CGS, see Fukasaku et al. \cite{fuk-iwa-sat2015} or references therein. 
    In this paper, the following notation is used. 
    Let $\succ$ be an admissible term order. For a polynomial $f\in K[\Abar,\Xbar]$ with
    a term order $\succ$ on $T(\Xbar)$, we regard $f$ as a polynomial in 
    $(K[\Abar])[\Xbar]$, which is the ring of polynomials with $\Xbar$ as variables and
    coefficients in $(K[\Abar])$ such that $\Abar$ is regarded as parameters.
    Given a term order $\succ$ on $T(\Xbar)$,
    $\lt(f)$, $\lc(f)$ and $\lm(f)$ denotes the leading term,
    the leading coefficient, and the leading monomial, respectively, satisfying that
    $\lt(f)=\lc(f)\lm(f)$ with $\lc(f)\in K[\Abar]$ and $\lm\in T(\Xbar)$
    (we follow the notation by Cox et al. \cite{cox-lit-osh2015}). 

    \begin{definition}[Algebraic partition and Segment]
        Let $S\subset C^m$ for $m\in\N$. A finite set 
        $\{\mathcal{S}_1,\dots,\mathcal{S}_t\}$ of nonempty subsets of $S$ 
        is called an algebraic partition of $S$ if it satisfies the following 
        properties:
        \begin{enumerate}
            \item $S=\bigcup_{k=1}^t\mathcal{S}_k$.
            \item For $k\ne j\in\{1,\dots,t\}$, $\mathcal{S}_k\cap\mathcal{S}_j=\emptyset$.
            \item For $k\in\{1,\dots,t\}$, $\mathcal{S}_k$ is expressed as
            $\mathcal{S}_k=V_C(I_1)\setminus V_C(I_2)$ for some ideals $I_1,I_2\subset K[\Abar]$.
        \end{enumerate}
        Furthermore, each $\mathcal{S}_k$ is called a segment.
    \end{definition}

    \begin{definition}[Comprehensive Gr\"obner System (CGS)]
        \label{def:CGS}
        Let $S\subset C^m$ and $\succ$ be a term order on $T(\Xbar)$. 
        For a finite subset $F\subset K[\Abar,\Xbar]$, a finite set
        $\mathcal{G}=\{(\mathcal{S}_1,G_1),\dots,(\mathcal{S}_t,G_t)\}$ is 
        called a Comprehensive Gr\"obner System (CGS) of $F$ over $\mathcal{S}$
        with parameters $\Abar$ with respect to $\succ$ if it satisfies
        the following:
        \begin{enumerate}
            \item For $k\in\{1,\dots,t\}$, $G_k$ is a finite subset of $K[\Abar,\Xbar]$.
            \item The set $\{\mathcal{S}_1,\dots,\mathcal{S}_t\}$ is an algebraic partition of 
            $\mathcal{S}$.
            \item For each $\cbar\in\mathcal{S}_k$, 
            $G_k(\bar{c},\bar{X})=\{g(\bar{c},\bar{X})\mid g(\Abar,\bar{X})\in G_k\}$
            is a Gr\"obner basis of the ideal 
            $\ideal{F(\bar{c},\bar{X})}\subset C[\bar{X}]$ with respect to $\succ$, where
            $F(\bar{c},\bar{X})=\{f(\bar{c},\bar{X})\mid f(\Abar,\bar{X}) \in F\}$.
            \item For each $\bar{c}\in S_k$, any $g\in G_k$ satisfies that 
            $\big(\lc(g)\big)(\bar{c})\neq 0$.
        \end{enumerate}
        Furthermore, if each $G_k(\bar{c},\bar{X})$ is a minimal or the reduced Gr\"obner basis,
        $\mathcal{G}$ is called a minimal or the reduced CGS, respectively. 
        In the case $\mathcal{S}=C^m$, the words ``over $\mathcal{S}$'' may be omitted.
    \end{definition}

    \subsection{Real root counting}
    \label{sec:real-root-counting}
    
    Let $I \subset {K}[\Xbar]$ be a zero dimensional ideal. Then,
    the quotient ring $K[\Xbar]/I$ is regarded as a finite dimensional vector
    space over $K$ \cite{cox-lit-osh2005}; let $\{v_1,\dots,v_d\}$ be its basis.
    For $h\in K[\Xbar]/I$ and $i,j$ satisfying $1\le i,j\le d$,
    let $\theta_{h,i,j}$ be a linear transformation defined as
    \[
        \begin{array}{cccc}
            {\theta_{h,i,j}} : & {K}[\bar{X}]/I & {\longrightarrow} & {K}[\bar{X}]/I \\
            & \rotatebox{90}{$\in$} & & \rotatebox{90}{$\in$} \\
            & f & \mapsto & hv_iv_jf
        \end{array}
        .
    \]
    Let $q_{h,i,j}$ be the trace of $\theta_{h,i,j}$ and $M_h^I$ be a symmetric matrix
    such that its $(i,j)$-th element is given by $q_{h,i,j}$.
    Let $\chi_h^I(X)$ be the characteristic polynomial of $M_h^I$, and $\sigma(M_h^I)$,
    called the signature of $M_h^I$, 
    be the number of positive eigenvalues of $M_h^I$ minus the number of negative 
    eigenvalues of $M_h^I$. Then, we have the following theorem on the real root counting
    \cite{bec-woe1994,ped-roy-szp1993}.

    \begin{theorem}[The real root counting theorem]
        We have
        \[
            \sigma(M_h^I)=\#(\{\cbar\in V_R(I)\mid h(\cbar)>0\})
            -\#(\{\cbar\in V_R(I)\mid h(\cbar)<0\}).
        \]
    \end{theorem}

    \begin{corollary}
        $\sigma(M_1^I)=\#(V_R(I))$.
    \end{corollary}

    Since we only consider a quantified formula with equations, as in 
    \Cref{eq:inverse-kinematic-qe-formula}, we omit properties of the real root 
    counting related to quantifier elimination of quantified formula with inequalities or inequations (for detail, see Fukasaku et al. \cite{fuk-iwa-sat2015}).

    \subsection{CGS-QE algorithm}
    \label{sec:cgs-qe-alg}

    The CGS-QE algorithm accepts the following quantified formula given as 
    \begin{gather*}
        \begin{array}[]{rccccc}
            \exists \Xbar 
            (f_1(\Abar,\Xbar)=0 & \land & \cdots & \land & f_\rho(\Abar,\Xbar)=0 & \land \\
             p_1(\Abar,\Xbar)>0 & \land & \cdots & \land & p_\sigma(\Abar,\Xbar)>0 & \land \\
             q_1(\Abar,\Xbar)\ne 0 & \land & \cdots & \land & q_\tau(\Abar,\Xbar)\ne 0),
        \end{array}
        \\
        \begin{split}
            f_1,\dots,f_\rho,p_1,\dots,p_\sigma,q_1,\dots,q_\tau &\in\Q[\Abar,\Xbar]\setminus\Q[\Abar],
        \end{split}
        \nonumber
    \end{gather*}
    then outputs an equivalent quantifier-free formula.
    Note that, in this paper, we give a quantified formula only with equations as shown 
    in \Cref{eq:inverse-kinematic-qe-formula}.
    The algorithm is divided into several algorithms. 
    The main algorithm is called \textbf{MainQE} and sub-algorithms are called
    \textbf{ZeroDimQE} and \textbf{NonZeroDimQE} for the case that
    the ideal generated by the component of the CGS is zero-dimensional or 
    positive dimensional, respectively.
    (For a complete description of the algorithm, see Fukasaku et al. \cite{fuk-iwa-sat2015}.)

    In the real root counting, we need to calculate $\sigma(M_h^I)$ as in 
    \Cref{sec:real-root-counting}. This calculation is executed using 
    the following property \cite{wei1998} derived from Descartes' rule of signs.
    Let $M$ be a real symmetric matrix of dimension $d$ and 
    $\chi(X)$ be the characteristic polynomial of $M$ of degree $d$,
    expressed as
    \[
        \begin{split}
            \chi(\lambda) &= \lambda^d+a_{d-1}\lambda^{d-1}+{\ldots}+a_0,\\
            \chi(-\lambda) &= (-1)^d\lambda^d+b_{d-1}\lambda^{d-1}+{\ldots}+b_0.
        \end{split}
    \]
    Note that $b_\ell=a_\ell$ if $\ell$ is even, and $b_\ell=-a_\ell$ if $\ell$
    is odd. 
    Let $\Lplus{\chi}$ and $\Lminus{\chi}$ be the sequence of the coefficients
    in $\chi(\lambda)$ and $\chi(-\lambda)$, defined as
    \begin{equation}
        \label{eq:L+-}
        \Lplus{\chi}=(1,a_{d-1},\dots,a_0),\quad
        \Lminus{\chi}=((-1)^d,b_{d-1},\dots,b_0),
    \end{equation}
    respectively.
    Furthermore, let $\Lbarplus{\chi}$ and $\Lbarminus{\chi}$ be the sequences
    defined by removing zero coefficients in $\Lplus{\chi}$ and
    $\Lminus{\chi}$, respectively, and let
    \begin{equation}
        \label{eq:S+-}
        \begin{split}
            \Splus{\chi} &= (\text{the number of sign changes in $\Lbarplus{\chi}$}), \\
            \Sminus{\chi} &= (\text{the number of sign changes in $\Lbarminus{\chi}$}).
        \end{split}
    \end{equation}
    Then, we have the following.

    \begin{lemma}
        Let $\Splus{\chi}$ and $\Sminus{\chi}$ be defined as in \Cref{eq:S+-}.
        Then, we have
        \[
                \Splus{\chi} = \#(\{c\in\R\mid c>0\land \chi(c)=0\}),\,
                \Sminus{\chi} = \#(\{c\in\R\mid c<0\land \chi(c)=0\}).
        \]
    \end{lemma}

    \begin{corollary}
        \label{cor:signature}
        Let $\Splus{\chi}$ and $\Sminus{\chi}$ be defined as in \Cref{eq:S+-},
        and
        $I$ be a zero dimensional ideal and
        $M_1^I$ be a matrix as in \Cref{sec:real-root-counting}.
        Then, we have 
        \begin{equation}
            \label{eq:signature}
            {\#}(V_{R}(I))=\sigma(M^I_1)=\Splus{\chi}-\Sminus{\chi}.
        \end{equation}
    \end{corollary}

    \subsection{Applying the CGS-QE algorithm for verification of real roots in the inverse kinematics problem}
    \label{sec:cgs-qe-alg-inverse-kinematics}

    We note that while the CGS-QE algorithm outputs an unquantified formula for the given 
    quantified formula, we apply the CGS-QE algorithm for verifying (or counting 
    the actual number of) real roots in the inverse kinematic problem, which means that 
    we may not necessarily derive an unquantified formula in the inverse kinematic 
    computation. 

    In the CGS-QE algorithm \cite{fuk-iwa-sat2015}, 
    for the quantified formula \eqref{eq:inverse-kinematic-qe-formula}, 
    \Cref{eq:signature} shows that
    \[
        {\#}(V_{R}(I))=\sigma(M^I_1)>0 \Leftrightarrow \Splus{\chi}\ne \Sminus{\chi}.
    \]
    In \Cref{eq:L+-}, by regarding $a_{d-1},\dots,a_0$ as variables, 
    $\Splus{\chi}\ne\Sminus{\chi}$ can be 
    expressed as a quantifier free first order formula, denoted as $I_d(a_0,\dots,a_{d-1})$.
    Then, \Cref{eq:inverse-kinematic-qe-formula} is expressed as an unquantified formula as
    \[
        (f_1=0 \land f_2=0 \land f_3=0 \land f_4=0 \land f_5=0 \land f_6=0) \land
        I_d(a_0,\dots,a_{d-1}).
    \]

    On the other hand, in verifying the existence of real roots in the inverse kinematic problem, 
    the coefficients of the characteristic polynomial $\chi(X)$ contain $x,y,z$,
    so do the elements in $\Lplus{\chi}$ and $\Lminus{\chi}$.
    Then, by substituting $x,y,z$ with the given coordinates of the end-effector, respectively,
    the values $\Splus{\chi}$ and $\Sminus{\chi}$ in \Cref{eq:S+-} are decided.
    Finally, with \Cref{eq:signature}, we calculate the number of real roots 
    for verifying the existence of real roots of the inverse kinematic problem.


    \section{Solving the inverse kinematic problem with the CGS-QE algorithm}
    \label{sec:inverse-kinematics-cgsqe}

    Let us consider the following inverse kinematic problem: 
    the manipulator consists of Joint $0,\dots,t+1$ with the coordinate system
    $\Sigma_0,\dots,\Sigma_{t+1}$, respectively, where
    Joint 0 represents the perpendicular foot on the ground from Joint 1 and
    Joint $t+1$ represents the end-effector as the example above.
    For $j=1,\dots,t$, let $\theta_j$ be the angle between axes $x_{j-1}$ and $x_j$
    with respect to the $z_j$ axis. 
    Let $(x,y,z)$ be the position of end-effector with respect to $\Sigma_0$,
    expressed as
    \begin{equation}
        \label{eq:cgsqe-inverse-kinematics}
        f_1(\bar{A},\bar{X})=0,\dots,f_w(\bar{A},\bar{X})=0,
    \end{equation}
    with $f_l(\bar{A},\bar{X})\in\Q[\bar{A},\bar{X}]$, 
    $\bar{A}=(x,y,z)$, $\bar{X}=(X_1,\dots,X_{2t})$,
    $X_{2j-1}=\cos\theta_j$, $X_{2j}=\sin\theta_j$ for $j=1,\dots,t$.
    
    Let $\bar{c}=(\alpha,\beta,\gamma)\in\R^3$ be 
    a position of the end-effector.
    For solving the inverse kinematic problem, we first verify the existence of 
    real roots in the polynomial equations
    \begin{equation}
        \label{eq:cgsqe-inverse-kinematics-equations}
        f_1(\bar{c},\bar{X})=0,\dots,f_w(\bar{c},\bar{X})=0,
    \end{equation}
    using the CGS-QE algorithm.
    If \Cref{eq:cgsqe-inverse-kinematics-equations} has real roots, 
    solve the system of equation numerically:
    let $(X_1,\dots,X_{2t})=(\psi_1,\varphi_1,\psi_2,\varphi_2,\dots,\psi_t,\varphi_{t})$
    be a solution.
    Then, a configuration of the joints is calculated as
    \begin{equation}
        \label{eq:cgsqe-inverse-kinematics-configuration}
        \theta_1=\arctan(\varphi_1/\psi_1),
        \theta_2=\arctan(\varphi_2/\psi_2),\ldots,\\
        \theta_t=\arctan(\varphi_{t}/\psi_{t}).
    \end{equation}
    
    Our method of solving the inverse kinematic problem consists of 
    \textit{preprocessing} steps as shown in \Cref{alg:preprocessing},
    followed by \textit{main} steps as shown in \Cref{alg:main}.
    Note that \Cref{alg:preprocessing} (the preprocessing steps) is executed for once 
    prior to solving the inverse kinematic problem, 
    then \Cref{alg:main} (the main steps) is executed every time for solving the 
    inverse kinematic problem.

    \begin{algorithm}[t]
        \caption{Preprocessing steps}
        \label{alg:preprocessing}
        \begin{algorithmic}[1]
            \Input{$f_1,\dots,f_w\in\Q[\Abar,\Xbar]$ in 
            \Cref{eq:cgsqe-inverse-kinematics-equations}, 
            $\succ$: a term order on $T(\bar{X})$};
            \Output{$\mathcal{G}=\{(\mathcal{S}_1,G_1,\chi_1(\lambda)),\dots,
            (\mathcal{S}_u,G_u,\chi_u(\lambda))\}$}
            \State{$\mathcal{G}\gets\emptyset$};
            \State{$\mathcal{F}=\{(\mathcal{S}_1,G_1),\dots,(\mathcal{S}_\mu,G_\mu)\}\gets
            \text{(a CGS of $\langle f_1,\dots,f_w\rangle$ with respect to $\succ$)}$};%
            \label{line:preprocessing:cgs}
            \For{$k=1,\dots,\mu$}
              \If{$\mathcal{S}_{k}\cap\R^3\ne\emptyset$ $\land$ 
              $\forall\cbar\in \mathcal{S}_{k}$ 
              ($G_{k}(\cbar,\Xbar)\ne\{0\}$  $\land$
              ($\langle G_{k}(\cbar,\Xbar)\rangle\ne \langle 1\rangle$))
              }%
              \label{line:preprocessing:if1}
                \If{($\forall\cbar\in\mathcal{S}_{k}$ $(\langle G_{k}(\cbar,\Xbar)\rangle$ is zero-dimensional)) }
                \label{line:preprocessing:if2}
                  \State{$M_1^{I_{k}}\gets\text{(the matrix defined as in 
                    \Cref{sec:real-root-counting} with $I_k=\langle G_{k}(\cbar,\Xbar)\rangle)$}$};%
                    \label{line:preprocessing:M}
                  \State{$\chi_{k}(\lambda)\gets\text{(the characteristic polynomial of $M_1^{I_{k}}$)}$};
                    \label{line:preprocessing:chi}
                  \State{$\mathcal{G}\gets\mathcal{G}\cup\{(\mathcal{S}_k,G_k,\chi_k(\lambda))\}$};
                \Else
                    \State{Let $\bar{A'}\subset\bar{A}$ and $\bar{X'}\subset\bar{X}$};
                    \State{For $G_k=(g_{k,1}(\bar{A},\bar{X}),\dots,g_{k,u_k}(\bar{A},\bar{X}))$, 
                     define 
                     \[
                     h_{k,1}(\bar{A'},\bar{X'}),\dots,h_{k,u'}(\bar{A'},\bar{X'})
                     \subset\Q[\bar{A'},\bar{X'}]
                     \]
                     \qquad\qquad\quad by setting appropriate constants to 
                     $\bar{A}\setminus\bar{A'}$ and $\bar{X}\setminus\bar{X'}$

                     \qquad\quad in 
                     $g_{k,1}(\bar{A},\bar{X}),\dots,g_{k,u_k}(\bar{A},\bar{X})$};
                    \State{$\mathcal{G}\gets\mathcal{G}\cup\text{(Output of Algorithm 1
                    with $h_{k,1}(\bar{A'},\bar{X'}),\dots,h_{k,u'}(\bar{A'},\bar{X'})$)}$};
                \EndIf
              \EndIf
            \EndFor
            \State{\Return $\mathcal{G}$};
        \end{algorithmic}
    \end{algorithm}

    \begin{algorithm}[bt]
        \caption{Main steps}
        \label{alg:main}
        \begin{algorithmic}[1]
            \Input{$\mathcal{G}=\{(\mathcal{S}_1,G_1,\chi_1(\lambda)),\dots,
            (\mathcal{S}_u,G_u,\chi_u(\lambda))\}$: the output of 
            \Cref{alg:preprocessing},
            with $G_k=(g_{k,1}(\bar{A},\bar{X}),\dots,g_{k,u_k}(\bar{A},\bar{X}))$;
            $\bar{c}=(\alpha,\beta,\gamma)\in\R^3$}: a position of the end-effector;
            \Output{($\theta_1,\dots,\theta_t$): a solution of 
            \Cref{eq:cgsqe-inverse-kinematics-equations} (if exists for 
            $\bar{c}=(\alpha,\beta,\gamma))$};
            \State{Find $k\in\{1,\dots,u\}$ satisfying that $\bar{c}\in\mathcal{S}_k$};
            \label{line:main:choosesegment}
            \If{$\nexists k\in\{1,\dots,u\}$ satisfying that $\bar{c}\in\mathcal{S}_k$} 
              \State{\Return $\emptyset$};
            \Else
                \State{$\bar{\chi}_k(\lambda)\gets\text{(set 
                $x\gets\alpha$, $y\gets\beta$, $z\gets\gamma$ in $\chi_{k}(\lambda)$)}$}; 
                \label{line:main:substitution}
                \State{Calculate $\Splus{\bar{\chi}_k}$, $\Sminus{\bar{\chi}_k}$ as in \Cref{eq:S+-}};
                \label{line:main:counting}
                \If{$\Splus{\bar{\chi}_k}-\Sminus{\bar{\chi}_k}=0$}
                    \Comment{\Cref{cor:signature}}
                    \State{\Return{$\emptyset$}};
                \Else
                    \Comment{There exists real roots of
                    \Cref{eq:cgsqe-inverse-kinematics-equations}}
                    \State{$                      
                            (\varphi_1,\psi_1,\dots,\varphi_{t},\psi_t)\gets\text{%
                            (a solution of $g_{k,1}(\bar{c},\Xbar)=\cdots=g_{k,u_k}(\bar {c},\Xbar)=0$%
                            )}$};\label{line:main:solve}
                    \For{$j=1,\dots,t$}
                        \State{$\theta_j\gets\arctan(\varphi_{j}/\psi_{j})$};
                        \label{line:main:theta}
                    \EndFor
                    \State{\Return{($\theta_1,\dots,\theta_t$)}};
              \EndIf
            \EndIf
        \end{algorithmic}
    \end{algorithm}

    In \Cref{alg:preprocessing}, the output $\mathcal{G}$ satisfies that each segment 
    $S_{k}$ in $\mathcal{G}$ contains real points and the system of polynomial equations  
    \eqref{eq:cgsqe-inverse-kinematics-equations} has finite number of roots
    for $\bar{c}\in S_{k}$. (Note that, at this point, the roots may not be real numbers.)
    Furthermore, \Cref{alg:preprocessing} follows the CGS-QE algorithm (for zero-dimensional case) in Fukasaku et al.\ \cite{fuk-iwa-sat2015} as follows.
    \begin{itemize}
        \item Line \ref{line:preprocessing:cgs} corresponds to line 2 in Algorithm 1 (MainQE).
        \item In line \ref{line:preprocessing:if1}, the condition $G_{k}(\cbar,\Xbar)\ne\{0\}$ corresponds to line 6 in Algorithm 1 (MainQE), and the condition 
        $\langle G_{k}(\cbar,\Xbar)\rangle\ne \langle 1\rangle$
        corresponds to line 1 in Algorithm 2 (ZeroDimQE). 
        \item Line \ref{line:preprocessing:if2} corresponds to line 9 in Algorithm 1 (MainQE). 
        \item Lines \ref{line:preprocessing:M} and \ref{line:preprocessing:chi} correspond to
         lines 5, 6, and 7 in Algorithm 2 (ZeroDimQE), and the result is used in verifying the existence of real roots for the given coordinates of the end-effector in 
         \Cref{alg:main}.
    \end{itemize}

    In line \ref{line:preprocessing:if2} of \Cref{alg:preprocessing}, in the case the ideal $\langle G_k(\bar{c}, \bar{X})\rangle$ is not zero-dimensional,
    the system of polynomial equations 
    $\{g(\bar{c},\bar{X})=0\mid g\in G_k(\bar{c},\bar{X})\}$
    satisfies that, for some variables, say 
    $\bar{X}'\subset\bar{X}$, the solution is not unique. 
    In this case, we set a solution to $\bar{X}'$,
    and, if necessary, choose a subset of parameters $\bar{A'}\subset\bar{A}$,
    then define polynomials
    $h_{k,j}(\bar{A'},\bar{X}')$, and call \Cref{alg:preprocessing} recursively.
    This computation is described as in lines 10--12, and our example is described in Section 4.2.

    In \Cref{alg:main}, for the given position of the end-effector $\bar{c}=(\alpha,\beta,\gamma)$, 
    we verify if there exist real roots of \Cref{eq:cgsqe-inverse-kinematics-equations}: 
    if there do, then, solve \Cref{eq:cgsqe-inverse-kinematics-equations} and calculate the 
    configuration of the joints.

    In the following, we explain our implementation\footnote{In the following footnotes,
    file locations are described as the relative path from the top directory of 
    the software repository \cite{snac2021}.} 
    \footnote{Note that, in the programs and the log files of Risa/Asir, index runs from 0 to $n-1$, while, in the results in the present paper and the notebook files of Mathematica, it runs from 1 to $n$.}  \cite{snac2021} and observations in computation details of the proposed algorithms for our example above.

    \subsection{\Cref{alg:preprocessing}}
    \label{sec:preprocessing-main}

    In \Cref{alg:preprocessing}, computation of CGS was executed with the computer algebra system 
    Risa/Asir and the rest of computation was executed by hand with Risa/Asir and Wolfram 
    Mathematica 12.0.0.

    \subsubsection{Computation of CGS}

    For $f_1,\dots,f_6\in\Q[\Abar,\Xbar]$ in
    \Cref{eq:inverse-kinematic-qe-formula},
    a CGS of the ideal $\langle f_1,\dots,f_6\rangle$ has been computed
    with respect to lexicographic order (``lexicographic'' 
    is abbreviated as the ``lex'' order)
    with 
    \begin{equation}
        \label{eq:variable-order}
        c_1\succ s_1\succ c_4\succ s_4\succ c_7\succ s_7
        \footnote{Since \Cref{eq:inverse-kinematic-qe-formula} is 
        fully existentially quantified formula and an order on the variables do not 
        affect the output, any order on the variables can be used.
        For future research direction, see \Cref{sec:remark}.}.  
    \end{equation}

    Computation of CGS was executed with the implementation of CGS computation by Nabeshima \cite{nab2018} on Risa/Asir.
    We have obtained a CGS with 
    the algebraic partition consisting of 33 segments in approximately 66.7 seconds using 
    the computing environment as shown in \Cref{sec:exp}.
    Let $\mathcal{F}$ be the computed CGS, expressed as
    \[
      \mathcal{F}=\{(\mathcal{S}_1,G_1),\dots,(\mathcal{S}_{33},G_{33})\}
      \footnote{The output code is available in Risa/Asir format as 
      \texttt{preprocessing-steps/cgs/F.rr} (text file, approximately 800 KB) and
      \texttt{preprocessing-steps/cgs/F.dat} (binary file, approximately 1 MB).
      Each $G_k$ is recorded as in \texttt{preprocessing-steps/cgs/F-basis/G-($k-1$).rr}.},
    \]
    with $\mathcal{S}_k=V_{\C}(I_{k,1})\setminus V_{\C}(I_{k,2})$,
    $I_{k,1}=\langle F_{k,1}\rangle$, $I_{k,2}=\langle F_{k,2}\rangle$%
    \footnote{For $\mathcal{S}_k$, $F_{k,1}$ and $F_{k,2}$ are
    recorded in the directory \texttt{preprocessing-steps/cgs/F-segments} as 
    \texttt{F-$(k-1)$-1.rr} and \texttt{F-$(k-1)$-2.rr}, respectively.}
    satisfying that $F_{k,1},F_{k,2}\subset\Q[x,y,z]$.


    \subsubsection{Verifying $\bm{\mathcal{S}_{k}\cap\R^3\ne\emptyset}$}

    We have verified the segments that contain real points as follows.
    \begin{enumerate}
        \item There exists a point in $V_{\R}(I_{k,1})$, explicitly found, which does not belong to $V_{\R}(I_{k,2})$
        (for $k=1,2,3$).
        \item With the discriminant or the QE computation, we see that there exists a point in 
        $V_{\R}(I_{k,1})$. Let $G_{k,1}$ be a Gr\"obner basis of $I_{k,1}$ with respect to lex order with $x\succ y\succ z$, and, for $f\in F_{k,2}$, the remainder of 
        $f$ divided by $G_{k,1}$ is not equal to zero (for $k=6,9,11,13,15,18,19,25,31,32$).
        The discriminant, the Gr\"obner basis and the QE computation were executed with Mathematica.
        \item There exists a point in $V_{\R}(I_{k,1})$ and $V_{\C}(I_{k,2})=\emptyset$ 
        (for $k=30$).
        \item We have $V_{\C}(I_{k,1})=\C^3$ and the QE computation shows that there exists a real point which is not contained in $V_{\C}(I_{k,2})$ (for $k=33$). The QE computation was 
        executed with Mathematica.
    \end{enumerate}
    
    Summarizing above\footnote{The computation was summarized as a Mathematica notebook file 
    \texttt{preprocessing-steps/cgs/F-segments/F-verification.nb}.}, we have $\mathcal{S}_{k}\cap\R^3\ne\emptyset$ for 
    \begin{equation}
        \label{eq:valid-segment-index}
        k\in\{1,2,3,6,9,11,13,15,18,19,25,30,31,32,33\}.
    \end{equation}

    \subsubsection{Verifying $\bm{G_k(\bar{c},\bar{X})\ne\{0\}}$ for $\bm{\bar{c}\in\mathcal{S}_k}$}

    For $k$ in \Cref{eq:valid-segment-index}, by Property 4 of \Cref{def:CGS}, $G_k$ satisfies 
    that $G_k(\bar{c},\bar{X})\ne\{0\}$ for $\bar{c}\in\mathcal{S}_k$.

    \subsubsection{Verifying $\bm{\langle G_{k}(\cbar,\Xbar)\rangle\ne\langle 1\rangle}$
     for $\bm{\bar{c}\in\mathcal{S}_k}$}

    For $k$ in \Cref{eq:valid-segment-index} satisfying $k\not\in\{3,30\}$, by Property 4 of \Cref{def:CGS}, $G_k$ satisfies that $\langle G_{k}(\cbar,\Xbar)\rangle\ne\langle 1\rangle$ for $\bar{c}\in\mathcal{S}_k$.

    \subsubsection{Verifying $\bm{\langle G_{k}(\cbar,\Xbar)\rangle}$ is zero-dimensional 
    for $\bm{\bar{c}\in\mathcal{S}_k}$}

    According to the ``Finiteness Theorem'' \cite[Chapter 5, Section 3, Theorem 6]{cox-lit-osh2015},
    for $k$ in \Cref{eq:valid-segment-index} satisfying $k\not\in\{3,30\}$, 
    $G_k$ satisfies that $\langle G_{k}(\cbar,\Xbar)\rangle$ is zero-dimensional for $\bar{c}\in\mathcal{S}_k$\footnote{The computation was executed with Risa/Asir with the program file \texttt{preprocessing-steps/cgs/zero-dimensional-test.rr} and the log file \texttt{preprocessing-steps/cgs/zero-dimensional-test.log}.\label{footnote:zero-dimensional}}. For $k\in\{3,30\}$, $(F_k,G_k)$ will be processed with 
    \Cref{alg:preprocessing} that is called recursively (see
    \Cref{sec:preprocessing-sub}).

    \subsubsection{Computation of $\bm{M_1^{I_{k}}}$ and $\bm{\chi_k(\lambda)}$}

    For $k$ in \Cref{eq:valid-segment-index} satisfying $k\not\in\{3,30\}$,
    the matrix $M_1^{I_{k}}$ and its characteristic polynomial 
    $\chi_{k}(\lambda)$ have been calculated with our implementation on Risa/Asir.
    Then, simplification of polynomial expressions on $\chi_{k}(\lambda)$ 
    has been executed with Mathematica. 
    For example, the output of Risa/Asir has a power of $2^{(1/2)}$ such as 
    $(2^{(1/2)})^4$ as in the form of \verb|(2^(1/2))^4|, thus,
    with Mathematica, we apply \texttt{Simplify} function to simplify $(2^{(1/2)})^4$ to $4$.

    \subsection{A recursive call of \Cref{alg:preprocessing}}
    \label{sec:preprocessing-sub}

    For $k\in\{3,30\}$, we see that $\langle G_{k}(\cbar,\Xbar)\rangle$ is not zero-dimensional because there does not exist $g\in G_k$ satisfying that $\lm(g)=s_1^{m_1}$ with $m_1>0$.
    Note that, for $k\in\{3,30\}$, $c_1^2+s_1^2-1\in G_k$. On the other hand, for $k\in\{3,30\}$,
    the points in $V_{\R}(I_{k,1})$ satisfy $x=y=0$. This means that the end-effector is located
    on the $z$-axis and $\theta_1$, the angle of Joint $J_1$, is not uniquely determined. 
    Thus, by setting $x=y=0$ and $\theta_1=0$ (i.e.\ $c_1=1,s_1=0$) in
    \Cref{eq:inverse-kinematic-equations}, we have 
    the following system of equations.
    \begin{equation}
        \label{eq:system-H}
        \begin{split}
           h_1 &= 112c_4s_7-16c_4-112s_4c_7+136s_4-44\sqrt{2}  = 0,\\
           h_2 &= 112c_4c_7+136c_4-112s_4s_7+16s_4+44\sqrt{2}+104-z=0,\\
           h_3 &= s_4^2+c_4^2-1=0,\quad h_4=s_7^2+c_7^2-1=0.
        \end{split}
    \end{equation}
    Note that, in \Cref{eq:system-H}, $f_2$ in \Cref{eq:inverse-kinematic-equations} vanishes by putting $s_1=0$, $f_4$ is eliminated, and $f_1,f_3,f_5,f_6$ are replaced with $h_1,h_2,h_3,h_4$,
    respectively.

    We have recursively applied \Cref{alg:preprocessing} to $\{h_1,h_2,h_3,h_4\}$ in 
    \Cref{eq:system-H}, as follows. The implementation used in each step is the same as the one 
    used in the corresponding step in the original call of \Cref{alg:preprocessing} as in above.
    
    \subsubsection{Computation of CGS}

    We have obtained a CGS with 
    the algebraic partition consisting of 3 segments in approximately 0.0142 seconds using 
    the computing environment as shown in \Cref{sec:exp}.
    Let $\mathcal{H}$ be the computed CGS, expressed as
    \[
        \mathcal{H}=\{(\mathcal{S}_{34},G_{34}),(\mathcal{S}_{35},G_{35}),(\mathcal{S}_{36},G_{36})\}
        \footnote{The output code is available in Risa/Asir format as 
        \texttt{preprocessing-steps/cgs/H.rr} (text file, approximately 1.8 KB) and
        \texttt{preprocessing-steps/cgs/H.dat} (binary file, approximately 7 KB).
        Each $G_k$ is recorded as in \texttt{preprocessing-steps/cgs/H-basis/G-($k-1$).rr}.},        
    \]
    with $\mathcal{S}_k=V_{\C}(I_{k,1})\setminus V_{\C}(I_{k,2})$,
    $I_{k,1}=\langle F_{k,1}\rangle$, $I_{k,2}=\langle F_{k,2}\rangle$%
    \footnote{For $\mathcal{S}_k$, $F_{k,1}$ and $F_{k,2}$ are
    recorded in the directory \texttt{preprocessing-steps/cgs/H-segments} as 
    \texttt{F-$(k-1)$-1.rr} and \texttt{F-$(k-1)$-2.rr}, respectively.}
    ($k=34,35,36$)
    satisfying that $F_{k,1},F_{k,2}\subset\Q[z]$.

    \subsubsection{Verifying conditions for $\bm{\mathcal{S}_k$}}
     We have $\mathcal{S}_{k}\cap\R^3\ne\emptyset$ only for $k=36$,
    with $V_{\C}(I_{36,1})=\C$ and the discriminant computation shows that none of the real point  
    is contained in $V_{\C}(I_{36,2})$\footnote{The computation was summarized as a 
    Mathematica notebook file \texttt{preprocessing-steps/cgs/H-segments/H-verification.nb}.}. 
    Furthermore, for $\bar{c}\in\mathcal{S}_{36}$, 
    $(\mathcal{S}_{36},G_{36})$ satisfies that 
    $G_{36}(\cbar,\Xbar)\ne\{0\}$, $\langle G_{36}(\cbar,\Xbar)\rangle\ne \langle 1\rangle$ and
    $\langle G_{36}(\cbar,\Xbar)\rangle$ is zero-dimensional\footnote{The verification of zero-dimensional is recorded in the same file as in Footnote~\ref{footnote:zero-dimensional}.}.
%


    \subsubsection{Computation of $\bm{M_1^{I_{k}}}$ and $\bm{\chi_k(\lambda)}$}

    For $(\mathcal{S}_{36},G_{36})$, The matrix $M_1^{I_{36}}$ and its characteristic polynomial 
    $\chi_{36}(\lambda)$ have been calculated, 
    and $\mathcal{G}=\{(\mathcal{S}_{36},G_{36},\chi_{36}(\lambda))\}$ has been returned to
    the original call of \Cref{alg:preprocessing}.

    \subsection{\Cref{alg:preprocessing} (continued)}

    Now we return to the original call of \Cref{alg:preprocessing} 
    as in \Cref{sec:preprocessing-main}.
    As a consequence, \Cref{alg:preprocessing} has returned the following output:
    \begin{equation}
        \label{eq:preprocessing-output}
        \begin{split}
            \mathcal{G} &=
            \{(S_{k},G_{k},\chi_k(\lambda))\mid k\in I'\}, \\
            I' &= \{1,2,6,9,11,13,15,18,19,25,31,32,33,36\}.
        \end{split}
    \end{equation}



    \subsection{\Cref{alg:main}}
    \label{sec:main}

    \Cref{alg:main} is implemented using SymPy on the top of Python and Risa/Asir
    connected with OpenXM infrastructure for communicating mathematical software systems, 
    as the implementation in our previous research \cite{hor-ter-mik2020}.
    A difference between our previous and present implementations is the purpose of 
    the use of Risa/Asir. While Risa/Asir has been used for computing Gr\"obner bases
    in our previous implementation, now it is used for the real root counting in the 
    present implementation.
    
    \subsubsection{Substituting $\bm{x,y,z}$ in
     $\bm{\chi_k(\lambda)}$ with $\bm{\bar{c}=(\alpha,\beta,\gamma)}$}

    For substituting $x,y,z$ in $\chi_k(\lambda)$,
    we first tried using Python for the entire step.
    However, a preliminary experiment had shown that it took 
    approximately 2 seconds for substituting $x,y,z$ in the coefficients for some 
    characteristic polynomials with the given position of 
    the end-effector $\bar{c}=(\alpha,\beta,\gamma)$ with SymPy, although 
    it took approximately 0.01 seconds to count 
    the number of sign changes of the sequence of coefficients, which was  
    sufficiently fast. Thus, we decided to use Risa/Asir for substituting $x,y,z$ 
    with $\bar{c}=(\alpha,\beta,\gamma)$, which is executed in approximately 
    0.2 seconds, and continue to use Python for counting 
    the number of sign changes of the sequence of coefficients.

    \subsubsection{Solving the system of polynomial equations}
    \label{sec:solvesystem}
    
    First, we remark on the form of the Gr\"obner bases 
    computed and re-organization of some of them. For $k\in I'\setminus\{18,25,31,32\}$, 
    where $I'$ is as shown in \Cref{eq:preprocessing-output}, 
    we see that the Gr\"obner basis $G_k=\{g_{k,1},\dots,g_{k,6}\}$ has a shape form such that
    \begin{align*}
        g_{k,1} &= g_{k,1}(s_7), \\
        g_{k,2} &= g_{k,2}(c_7,s_7), & \lm({g}_{k,2}) &= c_7, \\
        g_{k,3} &= g_{k,3}(s_4,c_7,s_7), &
        \lm(g_{k,3}) &= s_4, \\
        g_{k,4} &= g_{k,4}(c_4,s_4,c_7,s_7), & \lm(g_{k,4}) &= c_4, \\
        g_{k,5} &= g_{k,5}(s_1,c_7,s_7), &
        \lm(g_{k,5}) &= s_1, \\
        g_{k,6} &= g_{k,6}(c_1,s_1,c_7,s_7), & \lm(g_{k,6}) &= c_1.
    \end{align*}
    On the other hand, for $k\in\{18,25,31,32\}$, $G_k=\{g_{k,1},\dots,g_{k,7}\}$ with
    \begin{align*}
        g_{k,1} &= g_{k,1}(s_7), \\
        g_{k,2} &= g_{k,2}(c_7,s_7), & \lm(g_{k,2}) &= c_7s_7, \\
        g_{k,3} &= c_7^2+s_7^2-1, \\
        g_{k,4} &= g_{k,4}(s_4,c_7,s_7), & \lm(g_{k,4}) &= s_4, \\
        g_{k,5} &= g_{k,5}(c_4,s_4,c_7,s_7), &
        \lm(g_{k,5}) &= c_4, \\
        g_{k,6} &= g_{k,6}(s_1,c_7,s_7), & \lm(g_{k,6}) &= s_1, \\ 
        g_{k,7} &= g_{k,7}(c_1,s_1,c_7,s_7), & \lm(g_{k,7}) &= c_1.
    \end{align*}
    In the above formula, $g_{k,2}(c_7,s_7)$ has a term of $s_7^2$.
    Thus, let 
    $g'_{k,2}(c_7,s_7)$ be the result of substituting 
    the term of $s_7^2$ in $g_{k,2}(c_7,s_7)$ with $1-c_7^2$,
    then we have $\lm(g'_{k,2})=c_7^2$. 
    Furthermore, let $G'_i=\{g_{k,1},g'_{k,2},g_{k,4}\dots,g_{k,7}\}$.
    For example, in the case $i=18$, we have
    \begin{multline*}
        g_{18,2}  =
        - 1113167888 c_7 s_7
        - 1046586912 c_7
        + 4665483060 s_7^2\\
        + 123127872 s_7
        + 4239831888.
    \end{multline*}
    By substituting $s_7^2$ with $1-c_7^2$, we have
    \begin{multline*}
        g'_{18,2} =
        4665483060 c_7^2
        -1113167888 c_7 s_7
        - 1046586912 c_7\\
        + 123127872 s_7 
        -425651172,
    \end{multline*}
    with $\lm(g'_{18,2})=c_7^2$ \footnote{For $k=18,25,31,32$, computation of 
    $g'_{k,2}$ is saved in a Mathematica notebook file
    \texttt{main-steps/present-method/substitute-s7squared.nb}.}.
    
    As a result, for solving the system of polynomial equations,
    $G_k$ for $k\in I'\setminus\{18,25,31,32\}$ and 
    $G'_k$ for $k\in\{18,25,31,32\}$ are used\footnote{The check of shape form was 
    executed with Risa/Asir with the program file 
    \texttt{main-steps/present-method/shape-form-test.rr} and the log file
    \texttt{main-steps/present-method/shape-form-test.log}.}.

    The form of the Gr\"obner bases above shows that the system of polynomial
    equations can be solved by solving univariate equations successively, 
    as follows. 
    For $k\in I'\setminus\{18,25,31,32\}$, solve $g_{k,1}(s_7)=0$ and let $\varphi_7$ be 
    the calculated root. Next, solve $g_{k,2}(c_7,\varphi_7)=0$ for $c_7$ after 
    substituting $s_7$ with $\varphi_7$, and let $\psi_7$ be the calculated root.
    Then, solve $g_{k,3}(s_4,\psi_7,\varphi_7)=0$ for $s_4$ and
    $g_{k,5}(s_1,\psi_7,\varphi_7)=0$ for $s_1$
    after substituting $c_7$ and $s_7$ with $\psi_7$ and $\varphi_7$, respectively,
    and let $\varphi_4$ and $\varphi_1$ be the calculated roots, respectively.
    Finally, solve $g_{k,5}(c_4,\varphi_1,\psi_7,\varphi_7)=0$ for $c_4$ and 
    $g_{k,6}(c_1,\varphi_1,\psi_7,\varphi_7)=0$ for $c_1$ after
    substituting $s_1,s_4,c_7,s_7$ with $\varphi_1,\varphi_4,\psi_7,\varphi_7$, respectively,
    and let $\psi_4$ and $\psi_1$ be the calculated roots, respectively.
    For $k\in\{18,25,31,32\}$, the system of polynomial equations can be solved similarly.

    For solving the system of polynomial equations with Python, we use a numerical 
    solver in Python's NumPy package \cite{numpy2011} (\texttt{numpy.roots}) solving
    univariate equations, according to our observation in our previous research \cite{hor-ter-mik2020} showing that it was efficient and stable for solving systems of polynomial equations that were similar to the present ones.

    In the case more than one solution of the inverse kinematic problem is found,
    the program returns the first one in the solution set (see \Cref{sec:remark}).

    \section{Examples of the inverse kinematic computation}
    \label{sec:example}

    We demonstrate examples of the inverse kinematic computation\footnote{The examples were calculated with Risa/Asir with the program file \texttt{main-steps/present-method/cgs-qe-ik-example.py} and the log file \texttt{main-steps/present-method/log/cgs-qe-ik-example.log}.}.

\begin{figure}[t]
        \centering
        \includegraphics[scale=0.4]{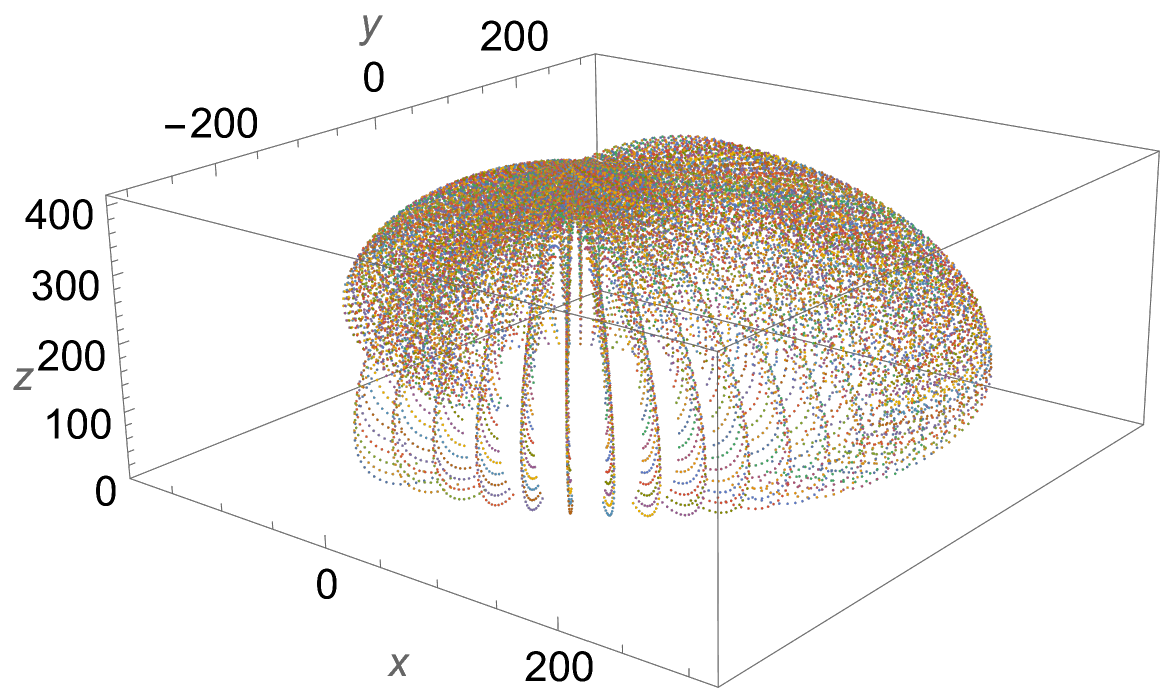}
        \caption{Positions of the end-effector of the EV3 manipulator for 
        $\theta_1,\theta_4,\theta_7\in\{-1.5,-1.4,\dots,1.5\}$.}
        \label{fig:feasible-region-1}
    \end{figure}

\begin{figure}[t]
        \includegraphics[scale=0.3]{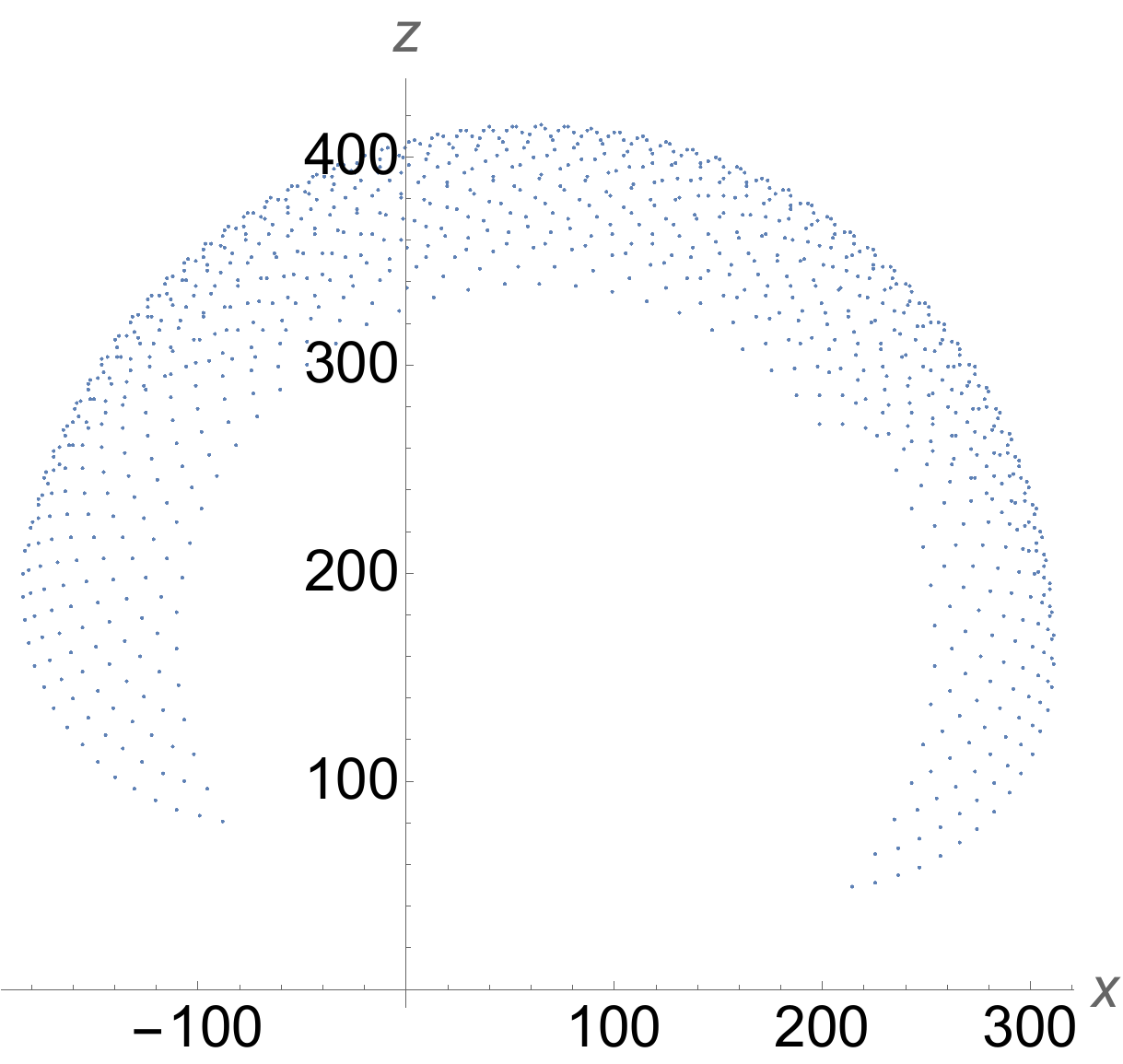}
        \caption{Positions of the end-effector of the EV3 manipulator for $\theta_1=0$, 
        $\theta_4,\theta_7\in\{-1.5,-1.4,\dots,1.5\}$.}
        \label{fig:feasible-region-2}
    \end{figure}

    To obtain ideas on the feasible region of the EV3 manipulator, 
    We have plotted positions of the end-effector for given configuration of the joints 
    with Mathematica\footnote{The computation is saved in a Mathematica notebook file
    \texttt{main-steps/present-method/ev3-feasible-region.nb}.}.
    \Cref{fig:feasible-region-1} shows the positions of the end-effector in the 
    $\R^3$ space for 
    $\theta_1,\theta_4,\theta_7\in\{-1.5(\simeq \pi/2),-1.4,\dots,1.5(\simeq\pi/2)\}$
    and \Cref{fig:feasible-region-2} shows them for $\theta_1=0$ and 
    $\theta_4,\theta_7\in\{-1.5,-1.4,\dots,1.5\}$ on the $xz$-plane.
    Note that the plotted points do not necessarily guarantee that they do not overlap
    with other components of the robot such as the pedestal or the Intelligent Brick: 
    in such a case, the range of the motion of the corresponding joint must be
    reduced.

    Since we have $V_{\C}(I_{33,1})=V_{\C}(\{0\})=\C^3$, we expect that most of the 
    given point $(x,y,z)\in\R^3$ belongs to segment $\mathcal{S}_{33}$. 
    For example, for $(x_1,y_1,z_1)=(-6061/41,-7679/51,4379/27)$, the solver 
    calculates that $(x_1,y_1,z_1)\in\mathcal{S}_{33}$. Then, the solver
    calculates that the system of polynomial equations has two real roots.
    By solving 
    the system of polynomial equations, the configuration of the joints is calculated as
    \begin{align*}
        & (\theta_1,\theta_4,\theta_7)\\
        &=(0.794578128292431-\pi,0.859415097289073-\pi,-1.38522249366716+\pi),\\
        &\qquad (0.794578128292431-\pi,-0.679494508722899,1.15100500453343-\pi).    
    \end{align*}
    
    Points in the feasible region which do not belong to segment $\mathcal{S}_{33}$ are
    those satisfying that $x=y=0$. 
    For example, for $(x_2,y_2,z_2)=(0,0,200)$, the solver 
    calculates that $(x_2,y_2,z_2)\in\mathcal{S}_{36}$.
    Then, the solver
    calculates that the system of polynomial equations has two real roots.
    By solving 
    the system of polynomial equations, the configuration of the joints is calculated as
    \begin{align*}
         (\theta_1,\theta_4,\theta_7)
        &=(0,0.236922524685754,-0.658765540873251+\pi),\\
        &\quad (0,-0.997268873826373+\pi,0.424548051739522-\pi).    
    \end{align*}

    For a point not in the feasible region, the solver first calculates if the point 
    belongs to a segment. If the point belongs to a segment, then the solver calculates 
    the number of real roots of the system of polynomial equations. If 
    the solver verifies that there exist real roots, it solves the system of polynomial equations.
    For example, for $(x_3,y_3,z_3)=(300,0,400)$, the solver 
    calculates that $(x_3,y_3,z_3)\in\mathcal{S}_{1}$.
    Then, the solver
    calculates that the system of polynomial equations has no real roots, and stops the computation.

    \section{Experiments}
    \label{sec:exp}

    We have tested our implementation along with a comparison of its performance
    with the previous one that we had proposed \cite{hor-ter-mik2020}, by solving inverse kinematic 
    problems with randomly given positions of the end-effector.

    Our experiments consist of 10 sets of tests conducted with 100 random positions 
    of the end-effector within the feasible region of the manipulator, given in 
    each set of experiments (thus, 1000 random points were given in total)\footnote{We have
    divided 1000 problems into 10 test sets, each with 100 problems because, in the 
    experiment of our previous work \cite{hor-ter-mik2020}, the computation of Risa/Asir
    stopped with an error when more than 100 problems were given at once.
    Although the cause of the error is not clear, since then, 
    we have divided problems into sets of 100 test cases to avoid such errors.}.
    The coordinates of the sample points are given as rational numbers with the magnitude of the denominator less than 100 (mm).
    
    The computing environment is as follows 
    (note that we have used a virtual machine on a desktop operating system).
    \begin{description}
        \item[Host environment] Intel Core i5-8259U CPU 2.30GHz, RAM 16GB, 
            macOS 11.4, Parallels Desktop for Mac Business Edition 16.5.1.
        \item[Guest environment] RAM 2GB, Linux 4.15.0, Python 3.6.9, 
            NumPy 1.19.5, SymPy 1.8, OpenXM 1.3.3, Risa/Asir 20210326 (Kobe Distribution). 
    \end{description}

    Note that, in comparison with our previous method, we have used the model of 
    the manipulator in the present paper.

    \begin{table}[t]
        \caption{A result of the inverse kinematics computation with real quantifier elimination.}
        \label{table:cgs-qe}
        \begin{center}
            \begin{tabular}{ccccc} \hline
                Test & $T_\textrm{Verify} $ (sec.) & $T_\textrm{Solve}$ (sec.) & $T_\textrm{Total}$ (sec.) & Error (mm) \\ \hline 
                $1$ & $0.367$ & $0.174$ & $0.541$ & $1.254{\times}10^{-9}$ \\ 
                $2$ & $0.365$ & $0.179$ & $0.544$ & $1.659{\times}10^{-8}$ \\ 
                $3$ & $0.370$ & $0.177$ & $0.548$ & $2.172{\times}10^{-9}$ \\ 
                $4$ & $0.369$ & $0.182$ & $0.551$ & $1.350{\times}10^{-9}$ \\ 
                $5$ & $0.367$ & $0.169$ & $0.536$ & $2.763{\times}10^{-9}$ \\ 
                $6$ & $0.367$ & $0.169$ & $0.536$ & $4.312{\times}10^{-8}$ \\ 
                $7$ & $0.367$ & $0.169$ & $0.536$ & $1.873{\times}10^{-9}$ \\ 
                $8$ & $0.367$ & $0.170$ & $0.536$ & $1.403{\times}10^{-9}$ \\ 
                $9$ & $0.367$ & $0.166$ & $0.533$ & $1.313{\times}10^{-9}$ \\ 
                $10$ & $0.368$ & $0.170$ & $0.538$ & $1.222{\times}10^{-8}$ \\ \hline 
                Average & $0.367$ & $0.172$ & $0.540$ & $8.405{\times}10^{-9}$ \\ \hline
            \end{tabular}
        \end{center}
    \end{table}
    \Cref{table:cgs-qe} shows the result of experiments of \Cref{alg:main}. 
    In each test, 
    $T_\textrm{Verify}$ is the sum of computing times of 
    Lines \ref{line:main:choosesegment}, \ref{line:main:substitution} and 
    \ref{line:main:counting},
    averaged over 100 examples, for verification of the existence of real roots.
    $T_\textrm{Solve}$ is the sum of computing times of 
    Lines \ref{line:main:solve} and \ref{line:main:theta},
    averaged over 100 examples, for solving a system of polynomial equations.
    $T_\textrm{Total}$ is the average of total computing time for inverse 
    kinematics computation,
    and `Error' is the average of the 
    absolute error, or the 2-norm distance of the end-effector from the randomly given position 
    to the calculated position with the configuration 
    of the computed joint angles $\theta_1,\theta_4,\theta_7$. 
    The bottom row `Average' shows the average values in each column of the 10 test sets.

    \begin{table}[t]
        \caption{A result of the inverse kinematics computation with our previous
        method \cite{hor-ter-mik2020}. Note that the model of the manipulator is renewed 
        as the one in the present paper.}
        \label{table:previous}
        \begin{center}
            \begin{tabular}{ccccc} 
                \hline
                Test & $T2_\textrm{GB}$ (sec.) & $T2_\textrm{Solve}$ (sec.) &
                 $T2_\textrm{Total}$ (sec.) & Error2 (mm) \\ 
                \hline
                $1$ & $0.497$ & $0.196$ & $0.693$ & $1.891{\times}10^{-9}$ \\
                $2$ & $0.472$ & $0.222$ & $0.694$ & $2.278{\times}10^{-9}$ \\
                $3$ & $0.474$ & $0.225$ & $0.699$ & $1.914{\times}10^{-9}$ \\
                $4$ & $0.451$ & $0.245$ & $0.696$ & $1.947{\times}10^{-9}$ \\
                $5$ & $0.497$ & $0.204$ & $0.701$ & $1.917{\times}10^{-9}$ \\
                $6$ & $0.473$ & $0.220$ & $0.693$ & $1.948{\times}10^{-9}$ \\
                $7$ & $0.495$ & $0.203$ & $0.699$ & $1.951{\times}10^{-9}$ \\
                $8$ & $0.474$ & $0.223$ & $0.698$ & $1.969{\times}10^{-9}$ \\
                $9$ & $0.479$ & $0.225$ & $0.704$ & $1.984{\times}10^{-9}$ \\
                $10$ & $0.476$ & $0.221$ & $0.697$ & $2.024{\times}10^{-9}$ \\
                \hline
                Average & $0.479$ & $0.219$ & $0.697$ & $1.982{\times}10^{-9}$ \\ 
                \hline
            \end{tabular}
        \end{center}
    \end{table}
    \Cref{table:previous} shows the result of experiments with our previous method
    for comparison, with the same inputs as those shown as in \Cref{table:cgs-qe}. 
    In each test, $T2_\textrm{GB}$ is the average computing time of
    Gr\"obner basis, $T2_\textrm{Solve}$ is the average computing time for
    solving the system of algebraic equations, 
    $T2_\textrm{Total}$ is the average of total computing time for inverse 
    kinematics computation,
    and `Error2' is the average of the absolute error, 
    or the 2-norm distance of the end-effector from the randomly given position 
    to the calculated position with the configuration 
    of the computed joint angles $\theta_1,\theta_4,\theta_7$.

    These results show that \Cref{alg:main} solves the inverse kinematic problem in a smaller amount of time than the previous method, 
    with verification of the existence of real roots.
    Also, in the present method, we have observed that the number of solutions of the 
    inverse kinematic problem was 2 or 4 for the given position of the end-effector. 
    (For further discussions, see \Cref{sec:remark}.)
    As for the accuracy of the solutions, since the actual size of the manipulator is 
    approximately 100 mm, computed solutions with the present method seem sufficiently accurate,
    although the average of errors in the present method 
    is slightly larger than that in the previous method.

    \section{Concluding remarks}
    \label{sec:remark}

    In this paper, we have presented a method and an implementation of 
    the inverse kinematics computation of a 3 DOF robot manipulator
    using the CGS-QE algorithm.
    Our method consists of 
    Algorithms \ref{alg:preprocessing} (the preprocessing steps) and
    \ref{alg:main} (the main steps).
    In \Cref{alg:preprocessing},
    using the CGS-QE algorithm, 
    we choose segments in the algebraic partition that have     
    real coordinates and calculate the characteristic polynomial of 
    the Hermite quadratic form.
    In \Cref{alg:main},
    for given parameters, after verification of the existence of 
    real roots using the results in \Cref{alg:preprocessing},
    a system of polynomial equations is solved after substituting the parameters in 
    a Gr\"obner basis in the CGS with the given values.

    Compared to our previous method, we see that the present method solves
    the inverse kinematic problem with the solutions as accurate as those solved 
    with the previous method. Furthermore, the present method has the following 
    benefits.
    \begin{enumerate}
        \item With verification of the existence of real roots using the CGS-QE algorithm, 
        one can judge the existence of the solution to the inverse kinematic problem. Furthermore, 
        if the given position of the end-effector is not feasible, one can avoid the useless computation of solving polynomial equations.
        \item The system of polynomial equations to be solved is constructed just by substituting parameters in the corresponding Gr\"obner basis with the given position of the 
        end-effector; it avoids the iterative computation of Gr\"obner basis for each given value.
    \end{enumerate}

    Experimental results show that \Cref{alg:main} in the present method solve the inverse 
    kinematic problem more efficiently than our previous method. 
    However, in the present method, one also needs time for executing \Cref{alg:preprocessing}.
    Thus, the previous method might be more efficient 
    for those who build an inverse kinematic solver from scratch for solving
    the inverse kinematic problem for just one choice of parameters.
    On the other hand, in other cases, such as using a pre-built solver 
    only with \Cref{alg:main}
    or solving the inverse kinematic problem (with \Cref{alg:preprocessing})
    for many choices of parameters, the present method 
    will be more desirable. Furthermore, it will be preferable to automate 
    the execution of \Cref{alg:preprocessing}, 
    and finding a threshold number of choice of parameters where the present method overwhelm 
    (including the time for executing \Cref{alg:preprocessing}) over the previous method 
    will be one of our next tasks.

    Other rooms for improvements on the present method or 
    future research directions include(s) the following.
    \begin{enumerate}
        \item If more than one solution of the inverse kinematic problem exists,
        we choose the first one in the list of solutions in our current implementation.
        However, it is desirable to choose an appropriate one based on certain criteria such 
        as the \emph{manipulability measure} \cite{sic-sci-vil-ori2008} which indicates how 
        the current configuration of the manipulator is away from a singular configuration.
        \item Experimental results show that the number of roots of the inverse kinematic 
        problem varies with the given position of the end-effector. Analyzing how 
        the number of roots changes in the feasible region will be preferable for choosing
        an appropriate one and understanding the problem's characteristics, such 
        as kinematic singularities.
        \item The quantified formula \eqref{eq:inverse-kinematic-qe-formula} we consider in this paper is 
        a fully existentially quantified formula, thus a satisfiable modulo theory (SMT) solver 
        such as Z3 \cite{mou-nik2008} might be useful for quantifier elimination.
        \item We still use Mathematica in \Cref{alg:preprocessing},
        despite our aim to use free software for building a solver. 
        Although it is no problem to use only \Cref{alg:main} with a pre-built solver 
        which incorporates only free software,
        building an implementation of \Cref{alg:preprocessing} with free computer algebra systems (such as SymPy or Z3 simplifier) is desired to achieve our goals.
        \item Although, our computation of CGS in \Cref{alg:preprocessing}
        (see \Cref{sec:cgs-qe}), an order of the variables was given 
        as in \Cref{eq:variable-order}, any other order can be used. It would be interesting to 
        observe how the change of order of the variables affects the result or the performance 
        of the algorithm.
        \item In the experiments, further investigation will be needed 
        to determine a reason for higher errors in the 
        solutions in the present method, although their magnitude seems sufficiently small  
        compared to the required accuracy of the solutions.
        We know that terms appearing in the system of polynomial equations in the present method 
        is different from those in the previous method: 
        in the previous method, the system of polynomial equations is given as
        \begin{align*}
            g_1(s_7) &= s_7^4 + r_1(s_7) = 0, & g_2(c_7,s_7) &= c_7 + r_2(s_7) = 0, \\
            g_3(s_4,s_7) &= s_4 + r_3(s_7) = 0, & g_4(c_4,s_7) &= c_4 + r_4(s_7) = 0, \\
            g_5(s_1,s_7) &= s_1 + r_5(s_7) = 0, & g_6(c_1,s_7) &= c_1 + r_6(s_7) = 0,
        \end{align*}
        where $r_i(s_7)$ ($i=1,\dots,6$) is a univariate polynomial in $s_7$ of degree 3.
        On the other hand, in the present method, the system of polynomial equations is given as in 
        \Cref{sec:solvesystem}, in which it seems that the polynomials are slightly complicated 
        (some polynomials have more number of variables in them) than those in the previous method,
        which might affect the errors in the solution.
    \end{enumerate}
    
    For our future research, applying the present method to inverse kinematics of 
    more complicated forms (such as more degree of freedom) might be interesting.
    Also, solving quantified formulas with inequality or inequation constraints is expected to 
    have a positive effect on the practicality of the present method.


\begin{thebibliography}{10}

        \bibitem{bec-woe1994}
        E.~Becker and T.~W\"{o}ermann.
        \newblock On the trace formula for quadratic forms.
        \newblock In {\em Recent advances in real algebraic geometry and quadratic
        forms ({B}erkeley, {CA}, 1990/1991; {S}an {F}rancisco, {CA}, 1991)}, volume
        155 of {\em Contemp. Math.}, pages 271--291. Amer. Math. Soc., Providence,
        RI, 1994.

        \bibitem{cha-mor-rou-wen2020}
        D.~Chablat, G.~Moroz, F.~Rouillier, and P.~Wenger.
        \newblock Using maple to analyse parallel robots.
        \newblock In J.~Gerhard and I.~Kotsireas, editors, {\em Maple in Mathematics
        Education and Research}, pages 50--64, Cham, 2020. Springer International
        Publishing.

        \bibitem{cox-lit-osh2005}
        D.~A. Cox, J.~Little, and D.~O’Shea.
        \newblock {\em {Using Algebraic Geometry}}.
        \newblock Springer, 2nd edition, 2005.

        \bibitem{cox-lit-osh2015}
        D.~A. Cox, J.~Little, and D.~O’Shea.
        \newblock {\em {Ideals, Varieties, and Algorithms: An Introduction to
        Computational Algebraic Geometry and Commutative Algebra}}.
        \newblock Springer, 4th edition, 2015.

        \bibitem{mou-nik2008}
        L.~de~Moura and N.~Bj{\o}rner.
        \newblock Z3: An efficient smt solver.
        \newblock In C.~R. Ramakrishnan and J.~Rehof, editors, {\em Tools and
        Algorithms for the Construction and Analysis of Systems}, pages 337--340.
        Springer, 2008.

        \bibitem{fau-mer-rou2006}
        J.-C. Faug{\`e}re, J.-P. Merlet, and F.~Rouillier.
        \newblock {On solving the direct kinematics problem for parallel robots}.
        \newblock Research Report RR-5923, {INRIA}, 2006.

        \bibitem{fuk-iwa-sat2015}
        R.~Fukasaku, H.~Iwane, and Y.~Sato.
        \newblock {Real Quantifier Elimination by Computation of Comprehensive
        Gr\"{o}bner Systems}.
        \newblock In {\em Proceedings of the 2015 ACM on International Symposium on
        Symbolic and Algebraic Computation}, ISSAC '15, page 173–180, New York, NY,
        USA, 2015. Association for Computing Machinery.

        \bibitem{hor-ter-mik2020}
        N.~Horigome, A.~Terui, and M.~Mikawa.
        \newblock {A Design and an Implementation of an Inverse Kinematics Computation
        in Robotics Using Gr{\"o}bner Bases}.
        \newblock In A.~M. Bigatti, J.~Carette, J.~H. Davenport, M.~Joswig, and
        T.~de~Wolff, editors, {\em Mathematical Software -- ICMS 2020}, pages 3--13,
        Cham, 2020. Springer International Publishing.

        \bibitem{kal-kal1993}
        C.~M. Kalker-Kalkman.
        \newblock {An implementation of Buchbergers' algorithm with applications to
        robotics}.
        \newblock {\em Mech. Mach. Theory}, 28(4):523--537, 1993.

        \bibitem{kap-sun-wan2010}
        D.~Kapur, Y.~Sun, and D.~Wang.
        \newblock A new algorithm for computing comprehensive gr\"{o}bner systems.
        \newblock In {\em Proceedings of the 2010 International Symposium on Symbolic
        and Algebraic Computation}, ISSAC '10, page 29–36, New York, NY, USA, 2010.
        Association for Computing Machinery.

        \bibitem{kaw-shi1999}
        H.~Kawasaki and T.~Shimizu.
        \newblock Development of robot symbolic analysis system: {ROSAM II} (in
        {J}apanese).
        \newblock {\em Journal of the Robotics Society of Japan}, 17(3):408--415, 1999.

        \bibitem{ros-complete}
        A.~Koubaa, editor.
        \newblock {\em {Robot Operating System (ROS): The Complete Reference}}, volume
        1--4.
        \newblock Springer, 2016--2020.

        \bibitem{mae-nor-oha-tak-tam2001}
        M.~Maekawa, M.~Noro, K.~Ohara, N.~Takayama, and K.~Tamura.
        \newblock The design and implementation of {OpenXM-RFC} 100 and 101.
        \newblock In K.~Shirayanagi and K.~Yokoyama, editors, {\em Computer
        Mathematics: Proceedings of the Fifth Asian Symposium on Computer Mathematics
        (ASCM 2001)}, pages 102--111. World Scientific, 2001.

        \bibitem{sympy2017}
        A.~Meurer, C.~P. Smith, M.~Paprocki, O.~{\v C}ert{\'i}k, S.~B. Kirpichev,
        M.~Rocklin, A.~Kumar, S.~Ivanov, J.~K. Moore, S.~Singh, T.~Rathnayake,
        S.~Vig, B.~E. Granger, R.~P. Muller, F.~Bonazzi, H.~Gupta, S.~Vats,
        F.~Johansson, F.~Pedregosa, M.~J. Curry, A.~R. Terrel, {\v S}.~Rou{\v c}ka,
        A.~Saboo, I.~Fernando, S.~Kulal, R.~Cimrman, and A.~Scopatz.
        \newblock {SymPy: symbolic computing in Python}.
        \newblock {\em PeerJ Computer Science}, 3, 2017.

        \bibitem{mon2018}
        A.~Montes.
        \newblock {\em The {Gr\"obner} Cover}.
        \newblock Springer, 2018.

        \bibitem{nab2018}
        K.~Nabeshima.
        \newblock {CGS}: a program for computing comprehensive {Gr\"obner} systems in a
        polynomial ring [computer software], 2018.
        \newblock \url{https://www.rs.tus.ac.jp/~nabeshima/softwares.html} (Accessed
        2021-10-24).

        \bibitem{net-spo1994}
        J.~Nethery and M.~Spong.
        \newblock {Robotica: a Mathematica package for robot analysis}.
        \newblock {\em IEEE Robotics \& Automation Magazine}, 1(1):13--20, 1994.

        \bibitem{nor2003}
        M.~Noro.
        \newblock A computer algebra system: {Risa/Asir}.
        \newblock In M.~Joswig and N.~Takayama, editors, {\em Algebra, Geometry and
        Software Systems}, pages 147--162. Springer, 2003.

        \bibitem{asir2018}
        {OpenXM Committers}.
        \newblock {Risa/Asir (Kobe Distribution) [computer software]}.
        \newblock \url{http://www.math.kobe-u.ac.jp/Asir/} (Accessed 2021-10-24).

        \bibitem{openxm}
        {OpenXM Committers}.
        \newblock {OpenXM}, a project to integrate mathematical software systems
        [computer software], 1998--2021.
        \newblock \url{http://www.openxm.org/} (Accessed 2021-10-24).

        \bibitem{ped-roy-szp1993}
        P.~Pedersen, M.-F. Roy, and A.~Szpirglas.
        \newblock Counting real zeros in the multivariate case.
        \newblock In {\em Computational algebraic geometry ({N}ice, 1992)}, volume 109
        of {\em Progr. Math.}, pages 203--224. Birkh\"{a}user Boston, Boston, MA,
        1993.

        \bibitem{pit-hil-ste-max-koc2008}
        J.~Pitt, D.~Hildenbrand, M.~Stelzer, and A.~Koch.
        \newblock Inverse kinematics of a humanoid robot based on conformal geometric
        algebra using optimized code generation.
        \newblock In {\em Humanoids 2008 --- 8th IEEE-RAS International Conference on
        Humanoid Robots}, pages 681--686, 2008.

        \bibitem{sic-kha2008}
        B.~Siciliano and O.~Khatib, editors.
        \newblock {\em Springer Handbook of Robotics}.
        \newblock Springer, 2008.

        \bibitem{sic-sci-vil-ori2008}
        B.~Siciliano, L.~Sciavicco, L.~Villani, and G.~Oriolo.
        \newblock {\em {Robotics: Modelling, Planning and Control}}.
        \newblock Springer, 2008.

        \bibitem{suz-sat2006}
        A.~Suzuki and Y.~Sato.
        \newblock A simple algorithm to compute comprehensive {Gr\"{o}bner} bases using
        {Gr\"{o}bner} bases.
        \newblock In {\em Proceedings of the 2006 International Symposium on Symbolic
        and Algebraic Computation}, ISSAC '06, page 326–331, New York, NY, USA,
        2006. Association for Computing Machinery.

        \bibitem{snac2021}
        A.~Terui, S.~Otaki, and M.~Mikawa.
        \newblock ev3-cgs-qe-ik: An inverse kinematics solver based on the {CGS-QE}
        algorithm for an {EV3} manipulator [computer software], 2021.
        \newblock \url{https://doi.org/10.5281/zenodo.5594896}.

        \bibitem{uch-mcp2011}
        T.~Uchida and J.~McPhee.
        \newblock {Triangularizing kinematic constraint equations using Gröbner bases
        for real-time dynamic simulation}.
        \newblock {\em Multibody System Dynamics}, 25:335--356, 2011.

        \bibitem{uch-mcp2012}
        T.~Uchida and J.~McPhee.
        \newblock {Using Gr\"obner bases to generate efficient kinematic solutions for
        the dynamic simulation of multi-loop mechanisms}.
        \newblock {\em Mech. Mach. Theory}, 52:144--157, 2012.

        \bibitem{numpy2011}
        S.~{van der Walt}, S.~C. {Colbert}, and G.~{Varoquaux}.
        \newblock {The NumPy Array: A Structure for Efficient Numerical Computation}.
        \newblock {\em Comput. Sci. Eng.}, 13(2):22--30, 2011.

        \bibitem{wal-sch2008}
        K.~Waldron and J.~Schmiedeler.
        \newblock Kinematics.
        \newblock in \cite{sic-kha2008}, pages 9--34.

        \bibitem{war2007}
        J.~Wall\'en.
        \newblock On robot modelling using {Maple}.
        \newblock Technical Report, Link\"oping University, 2007.

        \bibitem{wei1998}
        V.~Weispfenning.
        \newblock A new approach to quantifier elimination for real algebra.
        \newblock In B.~F. Caviness and J.~R. Johnson, editors, {\em Quantifier
        Elimination and Cylindrical Algebraic Decomposition}, pages 376--392, Vienna,
        1998. Springer Vienna.

    \end{thebibliography}

    \def\cprime{$'$}

\end{document}